\documentclass[10pt,twocolumn,letterpaper]{article}

\usepackage{iccv}
\usepackage{times}
\usepackage{epsfig}

\usepackage{graphicx}
\usepackage{amsmath}
\usepackage{amssymb}
\usepackage{booktabs}
\usepackage{multirow}
\usepackage{fixltx2e}
\usepackage{adjustbox}
\usepackage{tabularx}
\usepackage{caption}
\usepackage{subcaption}
\usepackage[pagebackref,breaklinks,colorlinks]{hyperref}
\usepackage{textcomp}
\usepackage[ruled,vlined]{algorithm2e}
\usepackage{algpseudocode}
\usepackage{algorithmicx}
\usepackage{algcompatible}

\usepackage{color}
\usepackage[table]{xcolor}
\usepackage{siunitx}
\usepackage{makecell,multirow}
\usepackage{graphicx}
\definecolor{turquoise}{cmyk}{0.65,0,0.1,0.3}
\definecolor{purple}{rgb}{0.65,0,0.65}
\definecolor{dark_green}{rgb}{0, 0.7, 0.35}
\definecolor{orange}{rgb}{0.7, 0.6, 0.1}
\definecolor{red}{rgb}{0.8, 0.2, 0.2}
\definecolor{darkred}{rgb}{0.6, 0.1, 0.05}
\definecolor{blueish}{rgb}{0.0, 0.3, .6}
\definecolor{light_gray}{rgb}{0.9, 0.9, 0.9}
\definecolor{pink}{rgb}{1, 0, 1}
\definecolor{greyblue}{rgb}{0.25, 0.25, 1}
\definecolor{amber}{rgb}{1.0, 0.75, 0.0}
\definecolor{aqua}{rgb}{0.0, 1.0, 0.9}
\definecolor{ballblue}{rgb}{0.13, 0.67, 0.8}

\newcommand\blank[1]{\rule[-.2ex]{#1}{.4pt}}

\newcommand{\paragrapht}[1]{\noindent\textbf{#1}}  

\newcommand{\method}{\textsc{CoT}\xspace}
\newcommand{\methodfull}{composition transformer\xspace}
\newcommand{\minus}{\text{-}}
 
\newcommand\blfootnote[1]{%
  \begingroup
  \renewcommand\thefootnote{}\footnote{#1}%
  \addtocounter{footnote}{-1}%
  \endgroup
}

\newcommand{\bs}{\mathbf{s}}
\newcommand{\mU}{\mathbf{U}}

\newcommand{\bp}{\mathbf{p}}
\newcommand{\mZ}{\mathbf{Z}}
\newcommand{\bz}{\mathbf{z}}
\newcommand{\mA}{\mathbf{A}}

\newlength\savewidth
\newcommand{\tablestyle}[2]{\setlength{\tabcolsep}{#1}\renewcommand{\arraystretch}{#2}\centering\footnotesize}
\renewcommand{\paragraph}[1]{\vspace{1.25mm}\noindent\textbf{#1}}

\newcolumntype{x}[1]{>{\centering\arraybackslash}p{#1pt}}
\newcolumntype{y}[1]{>{\raggedright\arraybackslash}p{#1pt}}
\newcolumntype{z}[1]{>{\raggedleft\arraybackslash}p{#1pt}}

\newcommand{\app}{\raise.17ex\hbox{$\scriptstyle\sim$}}

\newcommand{\x}{{\times}}
\definecolor{deemph}{gray}{0.6}

\definecolor{baselinecolor}{gray}{.9}

\usepackage[normalem]{ulem}
\usepackage[capitalize]{cleveref}
\crefname{section}{Sec.}{Secs.}
\Crefname{section}{Section}{Sections}
\Crefname{table}{Table}{Tables}
\crefname{table}{Tab.}{Tabs.}

\newcommand{\figref}[1]{Fig. \ref{#1}}
\newcommand{\tabref}[1]{Table \ref{#1}}
\newcommand{\equref}[1]{Eq. (\ref{#1})}

\iccvfinalcopy 

\def\iccvPaperID{8933} 

\ificcvfinal\fi

\begin{document}

\title{Hierarchical Visual Primitive Experts for Compositional Zero-Shot Learning}

\author{
Hanjae Kim$^{1}$\quad\quad 
Jiyoung Lee$^{2}$\quad\quad 
Seongheon Park$^{1}$\quad\quad 
Kwanghoon Sohn$^{1,3,}$\thanks{Corresponding author}
\vspace{5pt}\\
$^1$Yonsei University \quad\quad 
$^2$NAVER AI Lab \quad\quad 
$^3$Korea Institute of Science and Technology (KIST)\vspace{3pt}\\
{\tt\small{\{incohjk,sam121796,khsohn\}@yonsei.ac.kr}} \quad\quad\quad
\tt\small{lee.j@navercorp.com}
}
\vspace{5pt}

\maketitle
\ificcvfinal\thispagestyle{empty}\fi

\blfootnote{
This research was supported by the National Research Foundation of Korea (NRF) grant funded by the Korea government (MSIP) (NRF2021R1A2C2006703).
} 

\begin{abstract}
Compositional zero-shot learning (CZSL) aims to recognize unseen compositions with prior knowledge of known primitives (attribute and object).
Previous works for CZSL often suffer from grasping the contextuality between attribute and object, as well as the discriminability of visual features, and the long-tailed distribution of real-world compositional data. We propose a simple and scalable framework called Composition Transformer (CoT) to address these issues.
CoT employs object and attribute experts in distinctive manners to generate representative embeddings, using the visual network hierarchically. 
The object expert extracts representative object embeddings from the final layer in a bottom-up manner, while the attribute expert makes attribute embeddings in a top-down manner with a proposed object-guided attention module that models contextuality explicitly. 
To remedy biased prediction caused by imbalanced data distribution, we develop a simple minority attribute augmentation (MAA) that synthesizes virtual samples by mixing two images and oversampling minority attribute classes.
Our method achieves SoTA performance on several benchmarks, including MIT-States, C-GQA, and VAW-CZSL.
We also demonstrate the effectiveness of CoT in improving visual discrimination and addressing the model bias from the imbalanced data distribution.
The code is available at \url{https://github.com/HanjaeKim98/CoT}.

\vspace{-10pt}
\end{abstract}
\section{Introduction}
\label{sec:intro}

Humans perceive entities as hierarchies of parts; for example, we recognize `Cute Cat' by composing the meaning of `Cute' and `Cat'.
People can even perceive new concepts by composing the primitive meanings they already knew.
Such compositionality is a fundamental ability of humans for all cognition.
For this reason, compositional zero-shot learning (CZSL), recognizing a novel composition of known components (\ie, object and attribute), has been regarded as a crucial problem in the research community.

\begin{figure}[!t]
\centering
{\includegraphics[width=0.99\linewidth]{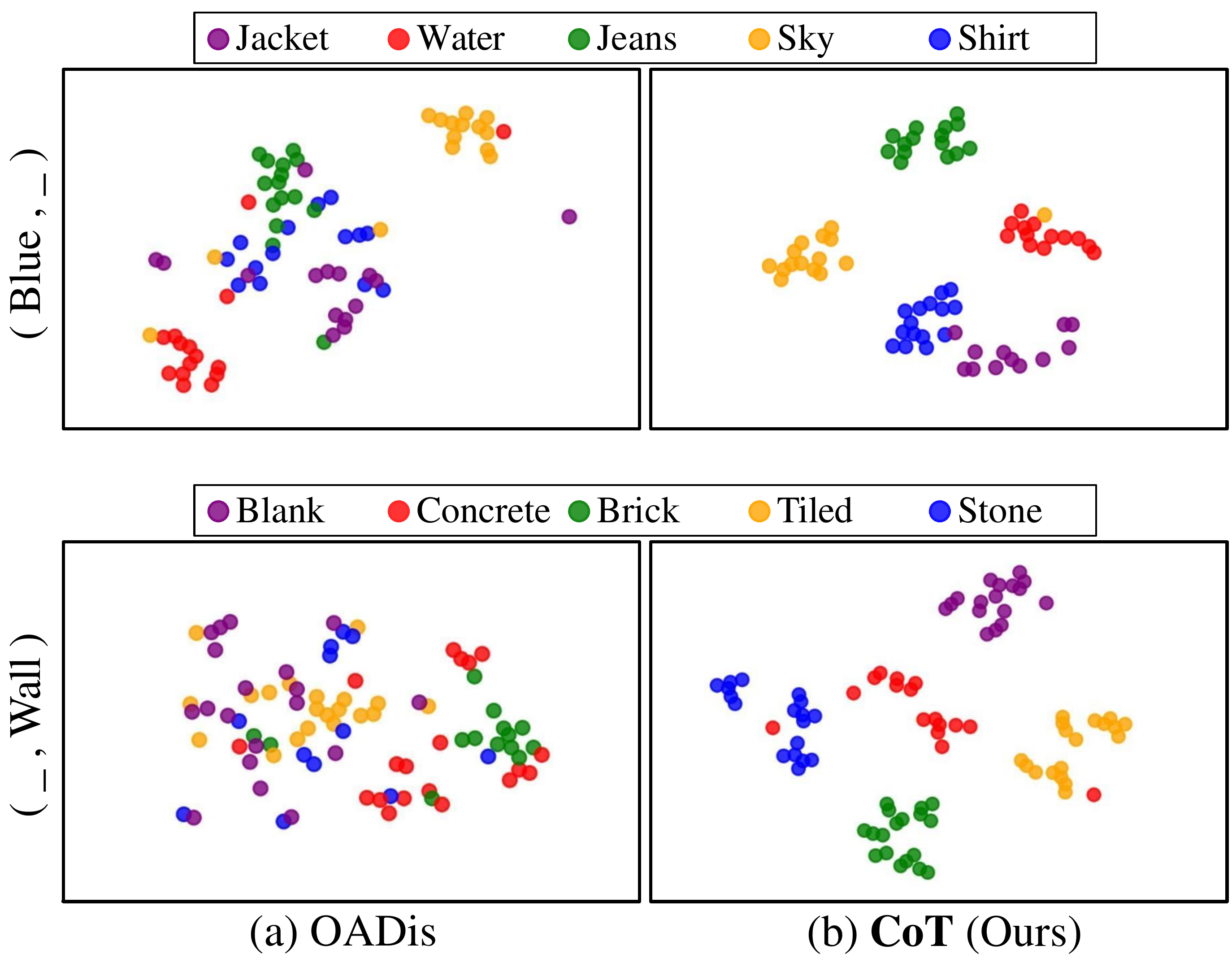}}\\ \vspace{-7pt}
\caption{Visual feature distribution of previous state-of-the-art approach (OADis)~\cite{saini2022disentangling} and \textbf{CoT} (Ours). The features are gathered with respect to attribute `Blue -' (Top) and object `- Wall' (Bottom). Notably, the previous method suffers from degraded visual discrimination such as (`Blue, Jeans' \vs `Blue, Shirt'), or all attributes composed into object `Wall'. 
}
\label{fig:1}
\end{figure}

A naive approach for CZSL is to combine attribute and object scores estimated from individually trained classifiers~\cite{misra2017red, karthik2021revisiting}. 
However, it is hard for these independent predictions to consider the interactions that result from a combination of different primitives, which is \textit{contextuality} (\eg different meaning of `Old' in `Old Car' and `Old Cat') in the composition~\cite{misra2017red}.
Previous approaches~\cite{nagarajan2018attributes, yang2020learning, mancini2021open, naeem2021learning, saini2022disentangling, zhang2022learning} have solved the problem by modeling the compositional label (text) embedding for each class.
They leverage external knowledge bases like Glove~\cite{mikolov2013efficient, pennington2014glove} to extract attribute and object semantic vectors, and concatenate them with a few layers into the composition label embedding.
The embeddings are then aligned with visual features in the shared embedding space, where the recognition of unseen image query becomes the nearest neighbor search problem~\cite{mancini2021open}. 
Nevertheless, the extent to which contextuality is taken into account in the visual domain has still been restricted.

Prior works~\cite{ruis2021independent, saini2022disentangling} have also pointed out the importance of \textit{discriminability} in visual features, which is related to generalization performance towards unseen composition recognition~\cite{tong2019hierarchical, min2020domain}.
A popular solution is disentanglement~\cite{ruis2021independent, saini2022disentangling, li2022siamese, zhang2022learning}, in which independent layers are allocated to extract intermediate visual representations of attribute and object. 
However, as shown in \figref{fig:1}, it may be challenging to extract their unique characteristics to understand the heterogeneous interaction.
We carefully argue that this problem arises from how the visual backbone (\eg ResNet18~\cite{he2016deep}) is used.
This is because attribute and object features, which require completely different characteristics~\cite{liang2018unifying, karthik2022kg}, are based on the same visual representation from the deep layer of the backbone. 

Another challenge in CZSL is the long-tailed distribution of real-world compositional data~\cite{asuncion2007uci, van2018inaturalist, atzmon2020causal}. Few attribute classes are dominantly associated with objects, which may cause a hubness problem~\cite{dinu2014improving, fei2021z} among visual features.
The visual features from the head (frequent) composition become a hub, which aggravates the visual features to be indistinguishable in the embedding space~\cite{fei2021z}, and induces a \textit{biased prediction} towards dominant composition~\cite{tang2020long,zhong2021improving}.

In this paper, based on the discussions above, we propose Composition Transformer (CoT) to enlarge the visual discrimination for robust CZSL.
Motivated by bottom-up and top-down attention~\cite{lin2017feature, liu2018path, li2022exploring}, the \method presents object and attribute experts, each forming its feature in different layers of the visual network.
Specifically, the object expert generates a representative object embedding from the last layer (\ie, bottom-up pathway) that is most robust to identify object category with high-level semantics~\cite{wu2019cascaded, pan2021scalable}.
Then we explicitly model contextuality through an object-guided attention module that explores intermediate layers of the backbone network and builds attribute-related features associated with the object (\ie, top-down pathway).
Based on this module, the attribute expert generates a distinctive attribute embedding in conjunction with the object embedding.
By utilizing all the features of each layer exhibiting different characteristics, our method comprehensively leverages a visual network to diversify the components' features in the shared embedding space.

Finally, we further develop a simple minority attribute augmentation (MAA) methodology tailored for CZSL to address the biased prediction caused by imbalanced data distribution~\cite{tang2020long}.
Unlike GAN-based augmentations~\cite{schonfeld2019generalized,li2022siamese} that lead to overwhelming computation during training, our method simplifies the synthesis process of the virtual sample by blending two images while oversampling minority attribute classes~\cite{chou2020remix, park2022majority}. 
Thanks to the label smoothing effect of the balanced data distribution by augmentation~\cite{zhang2017mixup, szegedy2016rethinking}, the \method is well generalized with minimal computational costs, resolving the bias problem induced by the majority classes.

Our contributions are summarized in three-folds:\vspace{-5pt}
\begin{itemize}
    \item To enhance visual discrimination and contextuality, we propose a Composition Transformer (\method). In this framework, object and attribute experts hierarchically estimate primitive embeddings by fully utilizing intermediate outputs of the visual network.\vspace{-5pt}
    \item We introduce a simple yet robust MAA that alleviates dominant prediction on head compositions with majority attributes.\vspace{-5pt}
    \item The remarkable experimental results demonstrate that our \method and MAA are harmonized to improve the performance on several CZSL benchmarks, showing state-of-the-art performance.
\end{itemize}
\section{Related Work}
\label{sec:related}

\begin{figure*}[!t]
\centering
{\includegraphics[width=0.99\linewidth]{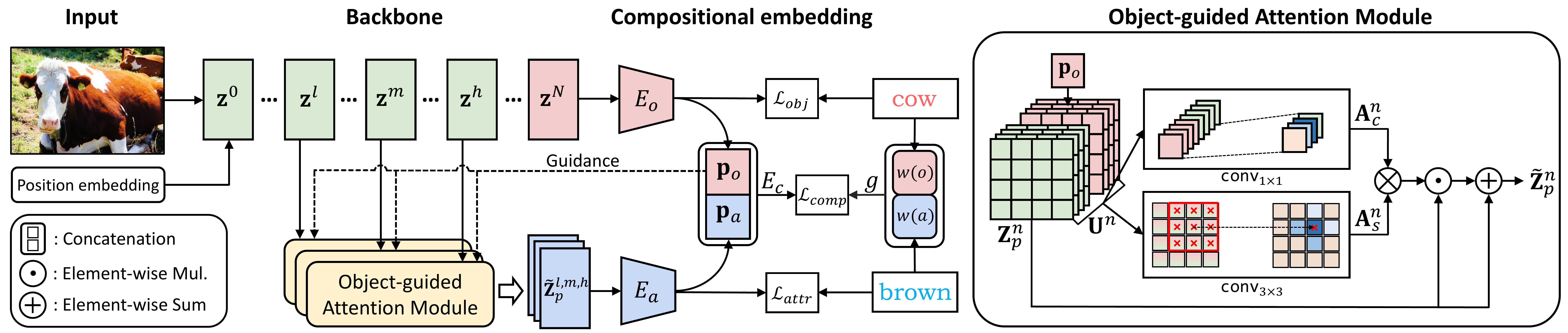}}\\ 
\vspace{-2pt}
\caption{The overall architecture of \methodfull. Object and attribute experts produce object- and attribute-specific features at different layers. An object-guided attention module is attached to each attribute experts to highlight attribute-specific features. Object-attribute embeddings are composed, and then projected to a joint embedding space with label embedding for optimization.}
\label{fig:2}
\vspace{-10pt}
\end{figure*}

Given a set of objects and their associated attributes, compositional zero-shot learning (CZSL) aims to recognize unseen compositions using primitive knowledge. 
Contrary to conventional methods~\cite{lu2016visual} that recognize the object and attribute independently, recent approaches~\cite{misra2017red,nagarajan2018attributes,li2020symmetry} have regarded contextuality as a crucial standpoint for understanding the interpretation of the attribute depending on the object class.
For example, AttrOps~\cite{nagarajan2018attributes} represented the attribute as a linear transformation of the object's state, and SymNets~\cite{li2020symmetry} regularized visual features to satisfy the group axioms with auxiliary losses.
Some works~\cite{naeem2021learning, xu2021relation} have leveraged graph networks to model the global dependencies among primitives and their compositions.

However, those methods have often suffered in distinguishing visual composition features.
Meta-learning~\cite{finn2017model} has been utilized for CZSL frameworks~\cite{purushwalkam2019task,wang2019tafe} to enhance the discriminative power of visual features but induce high training costs.
Other works~\cite{ruis2021independent,li2022siamese,saini2022disentangling} have disentangled representations of objects and attributes for modeling visual composition.
Those methods have shown promising generalization performance than conventional methods.
However, the aforementioned approaches have often failed to discriminate the minority classes in object and attribute, which is caused by severe model bias toward seen compositions~\cite{tang2020long, zhong2021improving}.
They have used the mostly-shared visual backbone for attribute and object representations, which needs more deep concern to be extracted differently according to its characteristics.
In specific, the attribute feature is changed in a wide variety of aspects as the property of the object changes, and the object feature must be perceived as semantically the same despite changing the attribute.

Our method is also closely related to visual attribute prediction \cite{patterson2016coco, krishna2017visual}, where the goal is to recognize visual attributes describing states or properties within an object. 
Since the fine-grained local information (\eg color, shape, and style) is required to the task, several approaches~\cite{sarafianos2018deep, tang2019improving, kalayeh2019symbiosis, shu2021learning, pham2021learning} utilize low-level features in backbone network, which contain textual details and objects' contour information~\cite{xie2021online, wei2021shallow}.
Meanwhile, with a similar motivation, some methods for CZSL have integrated low-level features with a multi-scale fusion approach to acquire robust visual features~\cite{wei2019adversarial,xu2021relation}. 
Although a simple integrated feature can recover spatial details slightly, it is hard to localize the RoI of the attribute that varies with the object for compositional recognition; for example, a puddle in an object `Street' for a composition class `Wet Street'.

Moreover, attributes in generic objects are long-tailed distributed in nature~\cite{sarafianos2018deep, pham2021learning}. 
The imbalance issue occurs as a biased recognition problem, because deep networks tend to be over-confident in head
composition classes \cite{samuel2021generalized, wallace2014improving}.
This phenomenon is also known as a hubness problem \cite{dinu2014improving, fei2021z}, where few visual features from head composition classes become indistinguishable from many tail composition samples.
One major solution for the data imbalance problem is the resampling~\cite{he2009learning, shen2016relay, geifman2017deep, mahajan2018exploring} by flattening the long-tailed dataset by over-sampling of minority classes or under-sampling of majority classes.
However, directly applying such sampling strategies in CZSL could intensify the overfitting problem on over-sampled classes, and discard valuable samples for learning the alignment between visual and semantic (linguistic) features in the joint embedding space~\cite{zhong2021improving}.
Another line of work is generation methods~\cite{wang2018low, wang2021rsg} that hallucinate a new sample from rare classes but they require complicated training strategies for data generation. In this work, we are inspired by recent data augmentation techniques~\cite{yun2019cutmix, chou2020remix, park2022majority, zhang2017mixup, chou2020adaptive}, and combine both resampling and generation approaches to tailor for CZSL, which facilitates the regularization of robust embedding space even in the minority classes via augmented virtual samples.

\section{Composition Transformer (\method)} 

In compositional zero-shot learning (CZSL), given an object set $\mathcal{O}$ and an attribute set $\mathcal{A}$, object-attribute pairs comprise a composition set $\mathcal{Y}$=$\mathcal{A} $$\times$$\mathcal{O}$.
The network is trained with an observable (seen) composition set $\mathcal{D}_{\text{tr}}$=$\{\mathcal{X}_{\text{s}}, \mathcal{Y}_{\text{s}}\}$, where $\mathcal{X}_{\text{s}}$ is an image set corresponding to a composition label set $\mathcal{Y}_{\text{s}}$.
Following the generalized zero-shot setting~\cite{purushwalkam2019task, chao2016empirical, xian2018zero}, a test set $\mathcal{D}_{\text{te}}$=$\{\mathcal{X}_{\text{te}}, \mathcal{Y}_{\text{te}}\}$ consists of samples drawn from both seen and unseen composition classes, \ie $\mathcal{Y}_{\text{te}}$=$\mathcal{Y}_{\text{s}}\cup \mathcal{Y}_{\text{u}}$ and $\mathcal{Y}_{\text{s}}\cap \mathcal{Y}_{\text{u}}$=$\varnothing$.
For this task, a powerful visual network is required to extract distinct features of objects and attributes respectively, in order to recognize both seen and unseen compositions well.

To this end, we propose a novel \methodfull (\method) that includes object and attribute expert modules as illustrated in~\figref{fig:2}. 
Our \method is basically built on the vision transformer (\ie ViT~\cite{dosovitskiy2020image}) and fully utilizes intermediate blocks to capture the attribute and object representations.
These representations are projected into the joint embedding space to calculate the similarity between given compositional labels and images.
We also introduce a minority attribute sampling method for data augmentation such that the biased recognition with head composition class is alleviated by label-smoothing regularization effect~\cite{muller2019does}. Next, we describe in detail the configurations of \method.

\subsection{Object expert}
The \method is composed of a sequence of transformer blocks including multi-head self-attention and MLP layers.
Generally, as the block becomes deeper, class discrimination capability is increased while being invariant from the attribute deformation~\cite{zeiler2014visualizing, naseer2021intriguing}.
Therefore, we select the final output from the last block to an input of the object expert in a bottom-up manner~\cite{lin2017feature, anderson2018bottom}. 

Formally, let us denote $\{\bz^n\}_{n=1}^N$ as the output features of the blocks in \method. The feature of $n$-th transformer blocks, $\bz^n$, consists of a class token (\ie, \texttt{[CLS]}) $\bz_c^n\in \mathbb{R}^{1\x D}$ and a set of patch tokens $\bz_p^n\in \mathbb{R}^{HW\x D}$.
The object expert $E_{o}$ encodes the last feature to present an object embedding: 
\begin{equation}
  \bp_o = E_{o}(\bz^{N}_c),
\end{equation}
where $E_{o}$ consists of a fully-connected (FC) layer.
To optimize the parameters in the object expert, we compute a cross-entropy loss with an object classifier whose weights are initialized from object word-embeddings as follows:
\begin{equation}
  \mathcal{L}_{\text{obj}} = - \log \frac{\exp \{  cos(\bp_o,w(o_{i})) /\tau_o  \} }{\sum_{o_{k} \in \mathcal{O}} \exp \{ cos(\bp_o,w(o_{k})) /\tau_o  \} } ,
  \label{eq:oloss}
\end{equation}
where $\tau$ is a temperature parameter, $w(o)$ denotes the word-embedding~\cite{pennington2014glove} corresponding to the object $o\in \mathcal{O}$.

\subsection{Attribute expert}

Capturing both attribute and object in the same deeply condensed feature is nontrivial as they require pretty different visual representations~\cite{liang2018unifying, karthik2022kg}. 
The straightforward solution is to design two-stream networks without any parameter sharing.
However, such individual training could not capture the contextuality~\cite{misra2017red}.
Hence, we build an attribute expert which utilizes semantics estimated from the object expert as a condition, and extracts attribute-related features on intermediate layers of visual backbone in a top-down fashion~\cite{lin2017feature, anderson2018bottom}.
We devise an object-guided attention module that leverages convolution kernel-based attention~\cite{woo2018cbam, chen2020image, kim2021prototype}, which excels at capturing fine-grained local information required for attribute recognition~\cite{huynh2020fine}.

Specifically, we first reshape $\bz^{n}_p$ to $\mZ^{n}_p$ with output resolution of $H\x W \x D$.
Following~\cite{ranftl2021vision}, we tile $\bp_o$ to match the spatial dimension with $\mZ^{n}_p$ and concatenate them to obtain an object-contextualized feature $\{\mU^{n}$=$[\bp_o,\mZ^{n}_p]\} \in \mathbb{R}^{H \x W \x 2D}$.
As shown in \figref{fig:2}, we generate a channel attention map $\mA^{n}_c \in \mathbb{R}^{1 \x 1 \x D}$ from the averaged feature across the \textit{spatial} axis:
\begin{equation}
    \mA_c^{n} = \sigma ( \texttt{conv}_{1 \x 1} (\texttt{avgpool}_{\text{spatial}}(\mU^{n})) ),
\end{equation}
where $\sigma$ denotes a sigmoid function.
We also generate a spatial attention map $\mA^{n}_s \in \mathbb{R}^{H \x W \x 1}$ from the averaged feature across the \textit{channel} axis:
\begin{equation}
    \mA^{n}_s = \sigma ( \texttt{conv}_{3 \x 3} (\texttt{avgpool}_{\text{channel}}(\mU^{n})) ).
\end{equation}
Finally, we aggregate two attention maps into a unified attention map $\mA^{n} \in \mathbb{R}^{H \x W \x D}$ as follows:
\begin{equation}
    \mA^{n} = \mA^{n}_s \times \mA^{n}_c.
\end{equation}
The output of our object-guided attention module is defined as attribute features with estimated attention maps, followed by the residual connection:
\begin{equation}
    \tilde{\mZ}^n_p = \mZ^n_p + \mA^{n} \odot \mZ^n_p,
\end{equation}
where $\odot$ is an operation for element-wise multiplication. 

Given $\tilde{\mZ}^n_p$ from each intermediate layer, we ensemble the features from different blocks by taking advantage of the hierarchical architecture~\cite{liu2021swin,naseer2021intriguing}.
It is required for recognizing heterogeneous types of attributes such as colors or shapes~\cite{pham2021learning}.
Especially, we choose low-, middle-, and high-level features from the backbone, denoted as $\{\tilde{\mZ}^l_p, \tilde{\mZ}^m_p, \tilde{\mZ}^h_p\}$, and ensemble those triplets with global average pooling (\texttt{GAP}) followed by concatenation.
The attribute expert $E_a$ then generates an attribute embedding with the triplets as
\begin{equation}
  \bp_{a} = E_{a}([\texttt{GAP}(\tilde{\mZ}^l_p), \texttt{GAP}(\tilde{\mZ}^m_p), \texttt{GAP}(\tilde{\mZ}^h_p)]),
     \label{eq:selfatt}
\end{equation}
where $E_{a}$ consist of a FC layer.
Similar to the object expert, we optimize the attribute expert with the cross-entropy loss: 
\begin{equation}
  \mathcal{L}_{\text{att}} = -\log \frac{\exp \{  cos(\bp_a,w(a_{i})) /\tau_a  \} }{\sum_{a_{k} \in \mathcal{A}} \exp \{ cos(\bp_a,w(a_{k})) /\tau_a  \} }.
  \label{eq:loss}
\end{equation}

\subsection{Mapping visual-semantic space}

The visual compositional embedding $\bp_x$ of the input image $x$ is calculated from object and attribute experts, formulated as
\begin{equation}
  \bp_x = E_{c}([\bp_o, \bp_a])).
\end{equation}
The embedding function $E_c$ includes an FC layer, projecting the visual features into the joint embedding space that aligns visual and semantic representations.
Following previous works~\cite{nagarajan2018attributes, mancini2021open}, we estimate semantic label embedding $\bp_y$ of each seen composition label $y \in \mathcal{Y}_{s}$ by projecting the concatenated word vectors into the joint space:
\begin{equation}
    \bs_y = g([w(o), w(a)]),
\end{equation}
where $g(\cdot)$ is a label embedding network that consists of 3 FC layers and ReLU activation function.
During training, we minimize the cosine distance of visual and semantic embeddings with cross-entropy loss as follows:
\begin{equation}
  \mathcal{L}_{\text{comp}}  =- \log \frac{\exp \{  cos(\bp_x,\bs_y) /\tau_c  \} }{\sum_{y_{k} \in \mathcal{Y}_{s}} \exp \{ cos(\bp_x,\bs_{y_k})) /\tau_c  \} },
  \label{eq:loss}
\end{equation}
where $\tau_c$ is a temperature parameter for the composition loss $\mathcal{L}_{\text{comp}}$.
Finally, the total objectives of \method is then formulated as follows:
\begin{equation}
  \mathcal{L}_{\text{total}} = \mathcal{L}_{\text{comp}} + \alpha \mathcal{L}_{\text{att}} + \beta \mathcal{L}_{\text{obj}}, 
  \label{eq:tloss}
\end{equation}
where $\alpha$ and $\beta$ are balance weights.

During inference, we measure the cosine similarity between a visual embeddings and all label embeddings from $\mathcal{Y}_{\text{te}}$ and regard it as a feasibility score of the image and composition labels~\cite{mancini2021open}. 
Following~\cite{zhang2022learning}, we also use the classification scores from object and attribute experts as a form of cosine similarity.
Therefore, a final feasibility score of label $y=(a,o)$ is derived by adding the above three scores, formulated as:
\begin{equation}
  c(y)  = cos(\bp_x, \bs_y) + cos(\bp_o,w(o)) + cos(\bp_a,w(a)),
  \label{eq:final}
\end{equation}
where we predict the label with the highest score as the final composition label.

\begin{figure}[t]
  \begin{subfigure}[b]{0.49\linewidth}
    \centering
    \includegraphics[width=\linewidth]{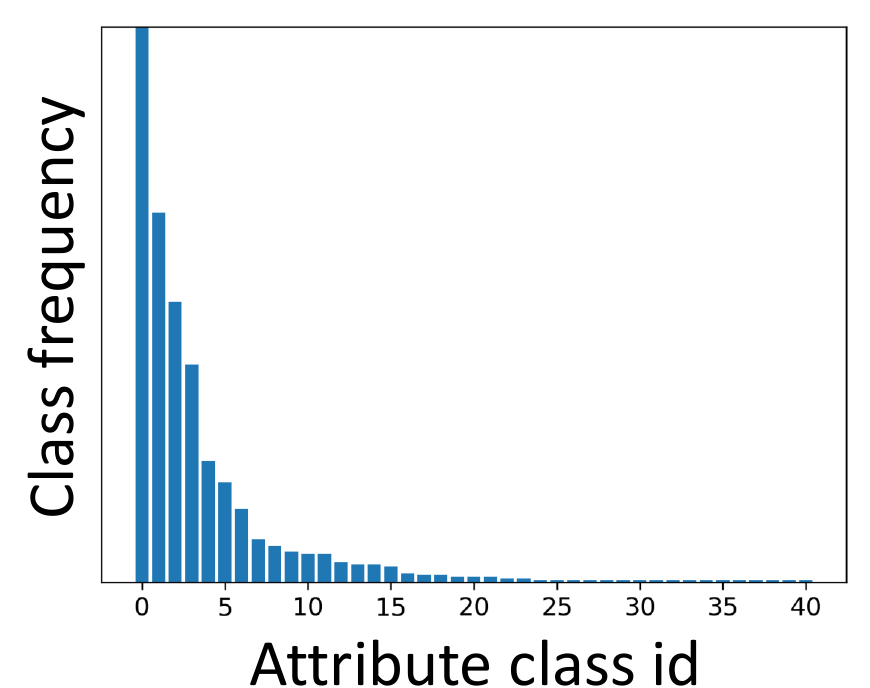}
    \caption{(\hspace{0.05cm} \blank{0.3cm}, Grass)}\label{fig:3a}
  \end{subfigure}
\hfill
  \begin{subfigure}[b]{0.49\linewidth}
    \centering
    \includegraphics[width=\linewidth]{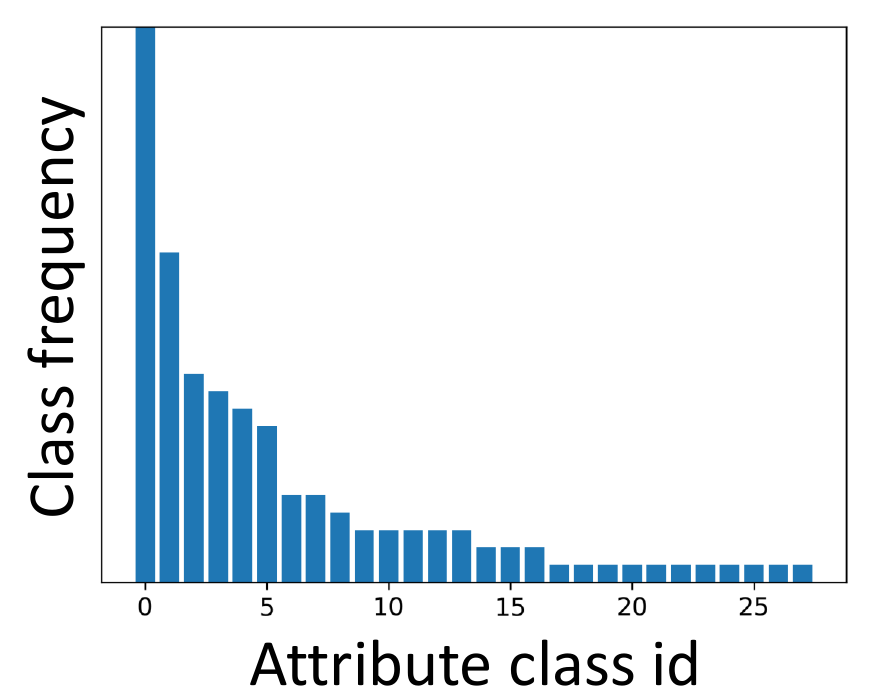}
 \caption{(\hspace{0.05cm} \blank{0.3cm}, Player)}\label{fig:3b}
  \end{subfigure}
    \vspace{-5pt}
  \caption{Attribute class distributions on VAW-CZSL. Every attributes in each plot are composed into a fixed object; (a) Grass and (b) Player.}\label{fig:3}
\end{figure}

\subsection{Minority attribute augmentation}
With a carefully designed architecture, our \method can generate distinctive visual features for CZSL.
Nevertheless, the imbalanced composition samples in the training dataset restrict the performance since attributes often co-occur with certain objects in the real world~\cite{liang2018unifying}.
For example in \figref{fig:3}, `Green Grass' is naturally more common than `White Grass'.
Such long-tailed data makes the deep models be over-confident in the composition with head attributes~\cite{zhong2021improving}, and thus hurts the generalization on unseen compositions.

To this end, we present a minority attribute augmentation (MAA) for training, which diversifies the minority composition.
First, two compositional embeddings from \method are selected as inputs to the augmentation, where $\bp_{x_{A}}$ and $\bp_{x_{B}}$ share the same object class.
Especially, to balance the major and minor attributes, we assign higher sampling weights on minority attributes with the weighted sampler~\cite{geifman2017deep}.
Formally, let $\zeta_{o_{i}}$ denote the number of attribute samples in the $i$-th attribute class composed on an object class $o$.
The sampling weight of the attribute-object pair $\kappa(a_i, o)$ is formulated by an inverse attribute class frequency according to:
\begin{equation}
  \kappa(a_i,o) = \frac{1 / \zeta_{o_{i}}}{\sum\nolimits_{i} 1 / \zeta_{o_{i}}}.
     \label{eq:sweight}
\end{equation}
Inspired by Mixup~\cite{zhang2017mixup, verma2019manifold}, we generate a virtual sample $\bp_{x_{M}}$ with alpha compositing and its attribute semantic embedding $w(a_{M})$ by linear combination:
\begin{align}
  \bp_{x_{M}} &= \lambda \bp_{x_{A}} + (1 - \lambda ) \bp_{x_{B}}, \nonumber \\
  w(a_{M}) &= \lambda w(a_A) + (1 - \lambda ) w(a_B).
\label{eq:vir}
\end{align}
The mixing factor $\lambda$ is sampled from a uniform distribution.

We notice that there are two significant differences from Mixup:
(1) Compared to Mixup sampling arbitrary images, MAA constructs a composition pair fastening an object with the minority aware sampling to address the imbalanced data issue in CZSL.  
(2) Mixup mixes one-hot vector labels \cite{zhang2017mixup, yun2019cutmix} for classification, while MAA generates a virtual label in word vector space, to jointly regularize visual and linguistic (label) embedding functions against the majority attribute classes with \equref{eq:tloss}.

\subsection{Extension to other backbones}
Our idea of \method is generic and scalable, so we can simply extend to other backbones such as ResNet-18 (R18)~\cite{he2016deep} by selecting specific layers from the backbone composed of multiple layers or blocks.
For a fair comparison to prior arts~\cite{naeem2021learning,saini2022disentangling,khan2022learning}, we study the impact of our contribution on the ResNet backbone based on the convolutional operation.
In specific, the object expert generates the object embedding from the last fifth convolution block.
The classification token is replaced by a global average pooled feature.
Similar to using ViT~\cite{dosovitskiy2020image}, we concatenate the intermediate layers into a feature vector for the attribute expert.
The remained settings such as the architecture of $E_{\text{comp}}$ and $g(\cdot)$ are the same as ViT.  
\vspace{-10pt}

\section{Experiments}
\subsection{Experimental setup}\label{sec:42}
\paragrapht{Datasets.}
MIT-States~\cite{isola2015discovering} dataset is a common dataset in CZSL that most previous methods adopted, containing 115 states , 245 object, and 1962 compositions.
C-GQA~\cite{naeem2021learning}  and VAW-CZSL~\cite{saini2022disentangling} are two brand-new and large-scale benchmarks collected from in-the-wild images with more general compositions, such as `Stained Wall' and `Hazy Mountain'.
Detailed data statistics including long-tailed distribution can be found in Appendix A.

\paragrapht{Metrics.}
We follow common CZSL setting~\cite{purushwalkam2019task}, testing performance on both seen and unseen pairs~\cite{chao2016empirical}.
Under the setting, we report area under the curve (AUC) ($\%$) between seen and unseen accuracies at different bias term in validation and test splits.
We also compute the best seen and unseen accuracies in the curve, and calculate the best harmonic mean to measure the trade-off between the two values.
Lastly, we follow~\cite{naeem2021learning} and report the attribute and object accuracies on unseen labels to measure the discrimination power of each representation.

\begin{table*}[t]

    \centering
    \scalebox{0.85}{
    \begin{tabular}{lcccccccccccccccc}
\toprule

\multirow{2}{*}{Methods} & \multirow{2}{*}{Backbone} 
 & \multicolumn{5}{c}{MIT-States} & \multicolumn{5}{c}{C-GQA} & \multicolumn{5}{c}{VAW-CZSL}  \\ 
 \cmidrule(lr){3-7}
 \cmidrule(lr){8-12}
  \cmidrule(lr){13-17}
  &  & V@1  & T@1 & HM & S & U & V@1  & T@1 & HM & S & U  & V@3  & T@3 & HM & S & U  \\ \midrule
  \multicolumn{5}{l}{\textbf{\textit{With frozen R18}}} \\
AoP~\cite{nagarajan2018attributes}       & R18 & 2.5  & 2.0  & 10.7  & 16.6 & 18.4  & 0.9  & 0.3  & 2.9  & 11.8 & 3.9   & 1.4 & 1.4 & 9.1 & 16.4 & 11.7  \\
LE+~\cite{misra2017red}      & R18 & 3.5 & 2.3 & 11.5 & 16.2 & 21.2 & 1.2  & 0.6  & 5.3  & 16.1 & 5.0  & 1.5 & 1.6 & 9.8 & 16.2 & 13.2  \\ 
TMN~\cite{purushwalkam2019task}     & R18 & 3.3 & 2.6 & 11.8 & 22.7 & 17.1 & 2.2  & 1.1  & 7.7  & 21.6 & 6.3   & 2.2 & 2.3 & 11.9 & 19.9 & 15.4  \\
Symnet~\cite{li2020symmetry}     & R18 & 4.5 & 3.4 & 13.8 & 24.8 & 20.0 & 3.3  & 1.8  & 9.8  & 25.2 & 9.2   & 2.3 & 2.3 & 12.2 & 19.1 & 15.8  \\
CompCos~\cite{mancini2021open}    & R18 & 6.9 & 4.8 & 16.9 & 26.9 & 24.5 & 3.5  & 2.6  & 12.1  & 28.1 & 11.8  & 3.1 & 3.2 & 14.2 & 23.9 & 18.0  \\
CGE*~\cite{naeem2021learning}      & R18 & 7.2 & 5.3 & 18.1 & 28.9 & 25.0 & 3.6  & 2.5  & 11.9  & 27.5 & 11.7   & 2.7 & 2.9 & 13.0 & 23.4 & 16.8 \\
SCEN~\cite{li2022siamese} & R18 & 7.2 & 5.3 & 18.4 & 29.9 & 25.2 & 4.0 & 2.9 & 12.4 & 28.9 & 12.1 & -  & - & - & - & - \\
OADis*~\cite{saini2022disentangling}     & R18 & \uline{7.6} & \uline{5.9} & 18.9 & \bf{31.1} & 25.6  & -  & -  & -  & - & -  & \uline{3.5} & \uline{3.6} & \uline{15.2} & \bf{24.9} & \uline{18.7} \\
CAPE~\cite{khan2022learning}  & R18 & - & 5.8 & \uline{19.1} & 30.5 & \uline{26.2} & - & \uline{4.2} & \uline{16.3} & \uline{32.9}  & \uline{15.6}  & - & - & - & - & - \\
\bf{\method}      & R18 & \bf{7.7} & \bf{6.2} & \bf{19.6} & \uline{30.8} & \bf{26.8} & \bf{4.9} & \bf{4.5} & \bf{16.6} & \bf{33.1} & \bf{16.6}   &  \bf{3.6} & \bf{3.8}  & \bf{15.7}  & \uline{24.6}  & \bf{19.1}  \\ 
\midrule
\multicolumn{5}{l}{\textbf{\textit{With frozen ViT-B}}} \\
CGE*~\cite{naeem2021learning}       & ViT-B & 8.7 & 7.3 & 21.3 & 33.5 & 28.6 & 4.5 & 3.8 & 15.6 & 31.3  & 14.4    & 3.7 & 3.9 & 15.7 & 26.3 & 19.7  \\
OADis*~\cite{saini2022disentangling}        & ViT-B & \uline{9.0} & \uline{7.5} & \uline{21.9} & \uline{34.2} & \uline{29.3}  & \uline{5.9}  & \uline{4.6}  & \uline{16.2}  & \uline{32.5} & \uline{15.3}  & \uline{4.1}  & \uline{4.3} & \uline{16.6} & \bf{26.9} & \uline{20.8}   \\ 
\bf{\method}      & ViT-B & \bf{9.2} & \bf{7.8} & \bf{23.2} & \bf{34.8} & \bf{31.5} & \bf{6.7}  & \bf{5.1}  & \bf{17.5}  & \bf{34.0} & \bf{18.8}  & \bf{4.4} & \bf{4.7} & \bf{17.7} & \bf{26.9} & \bf{22.2} \\ 
\midrule 
\multicolumn{5}{l}{\textbf{\textit{With finetuning ViT-B}}} \\
 CGE*~\cite{naeem2021learning}         & ViT-B & 11.4 & 9.7 & 24.8 & \bf{39.7} & 31.6 &  7.3 & 5.4  & 18.5  & 38.0 & 17.1  & 5.9 & 6.2 & 20.1 & 30.1 & 25.7 \\ 
 OADis*~\cite{saini2022disentangling}     & ViT-B & \uline{11.8} & \uline{10.1} & \uline{25.2} & 39.2 & \uline{32.1} & \uline{8.1}  & \uline{7.0}  & \uline{20.1}  & \uline{38.3} & \uline{19.8}   & \uline{6.2} & \uline{6.5} & \uline{20.4} & \uline{31.3} & \uline{26.1}  \\   
 \bf{\method}      & ViT-B & \bf{12.1} & \bf{10.5} & \bf{25.8} & \uline{39.5} & \bf{33.0} & \bf{8.7}  & \bf{7.4}  & \bf{22.1}  & \bf{39.2} & \bf{22.7}   & \bf{6.6} & \bf{7.2} & \bf{21.7} & \bf{32.9} & \bf{28.2}  \\
\bottomrule

\end{tabular}}\vspace{-5pt}
\caption{Quantitative comparison with prior arts. Following \cite{naeem2021learning, saini2022disentangling}, we measure AUC values on validation (V) and test (T) datasets, at top-1 (@1) on C-GQA and top-3 (@3) on VAW-CZSL, respectively. In addition, we report harmonic mean (HM), seen (S), unseen (US), attribute (A), and object (O) accuracies.
$*$ denotes the models accessing unseen composition labels at training.
The best result and the second best result are boldfaced and underlined, respectively.
}\label{tab:1} 
  
\end{table*}

\paragrapht{Baselines.}
We compare our \method with 9 recent CZSL approaches: AoP~\cite{nagarajan2018attributes}, LE+~\cite{misra2017red}, TMN~\cite{purushwalkam2019task}, Symnet~\cite{li2020symmetry}, CompCos~\cite{mancini2021open}, CGE~\cite{naeem2021learning}, SCEN~\cite{li2022siamese}, OADis~\cite{saini2022disentangling}, CAPE~\cite{khan2022learning}.
All baselines were built on pretrained ResNet-18 (R18) \cite{he2016deep}, with pre-trained word embedding such as GloVe~\cite{pennington2014glove}, FastText~\cite{bojanowski2017enriching}, and Word2Vec~\cite{mikolov2013efficient}.
To show the contribution of our proposed elements, we carry on the most recent methods including CGE and OADis with a transformer-based backbone (ViT-B)~\cite{dosovitskiy2020image}.
We also clarify that CGE and OADis utilize the knowledge of unseen composition labels at training for building a relation graph and hallucinating virtual samples respectively, while other methods do not assess the unseen composition label sets.
The results of other baselines were obtained from \cite{naeem2021learning, li2022siamese, khan2022learning, saini2022disentangling} or re-implemented from their official code.

\subsection{Implementation details}
\paragrapht{Models.}
We use ViT-B~\cite{dosovitskiy2020image} and ResNet18~\cite{he2016deep} pre-trained on ImageNet~\cite{deng2009imagenet} as a visual backbone.
In ViT-B backbone, Each patch size is $16 \times 16$, and the number of patch tokens in each layer is  $14 \times14 $, having $768$ channels per token.
The label embedding network $g$ is composed of 3 FC layers, two dropout layers with ReLU activation functions, where hidden dimensions are 900. 
We use one FC layer for $E_o$ and $E_a$ to produce 300-dimensional prototypical vectors that are matched with GloVe~\cite{pennington2014glove} embedding vectors from composition labels.
Therefore, both $E_{c}$ and $g(\cdot)$ produce 300-dimensional embedding vectors, with $l_2$ normalization.
In the attribute expert, we use the outputs of the 3rd, 6th and 9th blocks (denoted $\bz^{l}, \bz^{m}, \bz^{h}$ respectively) for multi-level feature fusion.
For ResNet18, we use the 2nd, 3rd and 4th blocks as $\bz^{l}, \bz^{m}, \bz^{h}$, resulting in a 438-dimensional feature vector.

\paragrapht{Training setup.}
We train \method with Adam optimizer \cite{kingma2014adam}.
The input images are augmented with random crop and horizontal flip, and finally resized into $224 \times 224$.
In ViT, learning rate is initialized to $1e\minus 4$ for all three datasets, and decayed by a factor of 0.1 for 10 epochs.
In R18, the initial learning rate is $5e\minus 5$ with the same decay parameter of the ViT setting.
Note that we freeze the GloVe~\cite{pennington2014glove} word embedding $w$ for fair comparisons.
For all datasets, we use the same temperature parameters~\cite{mancini2021open, saini2022disentangling, li2022siamese} of each cross-entropy loss, $\tau_c$, $\tau_o$, and $\tau_a$ as 0.05, 0.01 and 0.01 respectively.
We use the loss balance weights $\alpha$ and $\beta$ as (0.5, 0.5) on C-GQA, and (0.4, 0.6) on VAW-CZSL and MIT-States. 
We note that to prevent overfitting of \method we apply MAA from 15 and 30 epochs to ViT and R18, respectively.

\begin{table}[t]
    \small
    \centering
    \scalebox{0.99}{
    \begin{tabular}{lcccccc}
    \toprule
   \multirow{2}{*}{Methods}   & \multicolumn{2}{c}{MIT-States} & \multicolumn{2}{c}{C-GQA} & \multicolumn{2}{c}{VAW-CZSL}  \\ 
   \cmidrule(lr){2-3}
   \cmidrule(lr){4-5}
   \cmidrule(lr){6-7}

   &  Attr.  & Obj. & Attr.  & Obj. & Attr.  & Obj. \\ 
  \midrule
CGE~\cite{naeem2021learning}     &  \uline{35.7} &  44.4 & \uline{17.5}  & 34.6  & \uline{24.5} & 51.6 \\
OADis~\cite{saini2022disentangling}        &  35.2 & \uline{45.2}  & 14.7  & \bf{42.0} & 22.3 & \uline{53.8} \\
\bf{\method}     & \bf{37.3}  & \bf{46.0}  & \bf{19.8}  & \uline{40.2} & \bf{27.3}  & \bf{54.2} \\

\bottomrule

\end{tabular}}
\vspace{-.4em}
\caption{Comparison of unseen attribute (Attr.) and object (Obj.) occuracies with the latest CZSL methods~\cite{naeem2021learning, saini2022disentangling}. All results are reported on fine-tuned ViT backbone.}\label{tab:2}
\end{table}

\begin{table*}[t]
\vspace{-.2em}
\centering
\subfloat[Component analysis.
\label{tab:3a}
]{
\centering
\begin{minipage}{0.26\linewidth}{\begin{center}
\tablestyle{4pt}{1.05}
    \begin{tabular}{ccccc}
    \toprule
    CoT & MAA &  AUC & HM \\ 
    \midrule
     & & 5.81 & 18.1 \\ 
     & \checkmark & 6.05 & 19.0 \\
    \checkmark &  & 6.93 & 20.8 \\
    \checkmark & \checkmark &  \bf{7.20} & \bf{21.7} \\
\bottomrule
\end{tabular}
\end{center}}\end{minipage}
}
\subfloat[
Impact of loss functions.
\label{tab:3b}
]{
\begin{minipage}{0.26\linewidth}{\begin{center}
\tablestyle{4pt}{1.05}
    \begin{tabular}{lcc}
    \toprule
    Loss component & AUC & HM \\ 
    \midrule
    $\mathcal{L}_{\text{comp}}$ & 5.94 & 18.9 \\
    $\mathcal{L}_{\text{comp}}$+$\mathcal{L}_{\text{obj}}$ & 6.22 & 19.6 \\
    $\mathcal{L}_{\text{comp}}$+$\mathcal{L}_{\text{attr}}$ & 6.91 & 21.4 \\
    $\mathcal{L}_{\text{comp}}$+$\mathcal{L}_{\text{attr}}$+$\mathcal{L}_{\text{obj}}$ & \bf{7.20} & \bf{21.7} \\
\bottomrule
\end{tabular}
\end{center}}\end{minipage}
}\hspace{8pt}
\subfloat[
Impact of data augmentation.
\label{tab:3c}
]{
\begin{minipage}{0.25\linewidth}{\begin{center}
\tablestyle{3pt}{1.05}
    \begin{tabular}{lcc}
    \toprule
    Augmentation & AUC & HM \\ 
    \midrule
    Manifold Mixup~\cite{verma2019manifold} & 5.93 & 18.8 \\
    CutMix~\cite{yun2019cutmix} & 6.95 & 20.7 \\
    Mixup~\cite{zhang2017mixup} & 7.04 & 21.0 \\
    MAA & \bf{7.20} & \bf{21.7} \\
\bottomrule
\end{tabular}
\end{center}}\end{minipage}
}

\subfloat[
Impact of object guidance.
\label{tab:3d}
]{\centering
\begin{minipage}{0.42\linewidth}{\begin{center}
\tablestyle{3pt}{1.05}
    \begin{tabular}{lcccccc}
    \toprule
     Guidance & AUC & HM & S & U & A & O\\
     \midrule
     w/o guidance & 6.7 & 20.5 & 30.7 & 25.8 & 22.5 & 53.4 \\
     Attribute guidance & 6.9 & 21.0 & 31.5 & 26.4 & 25.7 & 53.9 \\
     Object guidance & \bf{7.2} & \bf{21.7} & \bf{32.9} & \bf{28.2} & \bf{27.3} & \bf{54.2} \\
\bottomrule
\end{tabular}
\end{center}}\end{minipage}
}
\subfloat[
Different configurations of intermediate blocks in attribute experts.
\label{tab:3e}
]{
\begin{minipage}{0.45\linewidth}{\begin{center}
\tablestyle{3pt}{1.05}
\setlength{\tabcolsep}{4pt}
    \begin{tabular}{lccccccccc}
    \toprule

    Ensemble & $z_l$ & $z_m$ & $z_h$ &  AUC &  HM &  S & U  & A  & O  \\
      
    \midrule
    Low & 2 & 3 & 4 & 6.6 & 20.7 & 30.3 & 28.7 & \uline{25.8} & 52.3 \\ 
    Mid & 5 & 6 & 7 & \uline{6.9} & 21.2 & 31.9 & \bf{29.7} & 25.2 & \bf{55.9} \\
    High & 9 & 10 & 11 & 6.8 & \uline{21.4} & \bf{33.4} & 26.6 & 24.1 & \uline{54.5} \\
    Mixture & 3 & 6 & 9 & \bf{7.2} & \bf{21.7} & \uline{32.9} & \uline{28.2} & \bf{27.3} & 54.2 \\

\bottomrule
\end{tabular}
\end{center}}\end{minipage}
}
\centering
\\

\vspace{-.4em}
\caption{Ablation experiments on VAW-CZSL dataset. Note that all results are reported on \method (ViT-B), with a fine-tuning setting.}
\label{tab:3} 
\end{table*}

\subsection{Main evaluation}
\tabref{tab:1} compares generalization performance between our methods and the baselines with ResNet18~\cite{he2016deep} (R18) and ViT-B~\cite{dosovitskiy2020image} backbone setting.
To further compare visual discrimination, we also compare unseen attribute and object accuracies of our methods and the latest CZSL methods~\cite{naeem2021learning, saini2022disentangling} in \tabref{tab:2}.
In the following, we analyze the results from the three datasets.

\paragrapht{MIT-States.}
From \tabref{tab:1}, the \method achieves the best AUC values (V@1 and H@1) and harmonic mean (HM) with both R18 and ViT backbone.
 These results sufficiently demonstrate our experts’ applicability on the CNN-based backbone.
With a finetuned ViT-B, the \method obtains the best validation AUC of $12.1\%$ and test AUC  of $10.5\%$, surpassing previous CZSL methods with a large margin.
Our method also improves attribute (Attr.) and object (Obj.) accuracies as in \tabref{tab:2}, showing the effectiveness of \method for visual discrimination.

\paragrapht{C-GQA.}
Similar trends can be observed in the C-GQA dataset.
In the frozen R18 setting, the \method achieves the best test AUC (T@1) of $4.5\%$, comparable with previous state-of-the-art CAPE.
Although the T@1 scroes from CGE and OADis are also increased upon replacing the backbone from R18 to ViT-B, our \method with ViT-B outperforms OADis with larger margins in all metrics, as compared to the performance with R18.
It is noticeable that our proposed augmentation technique contributes the unbiased prediction, improving unseen accuracies into $18.8\%$ with a minority-aware sampling.
The performance regarding AUC is improved with end-to-end training with backbone, $2.0\%$ in V@1 and $2.3\%$ in T@1.
Of note, OADis and CGE access unseen composition labels at training, so it is unfair to directly compare with other methods including \method.
Nevertheless, more surprisingly, our \method shows comparable and superior performance to OADis.
Finally, in \tabref{tab:2}, our model also improves the attribute accuracy by achieving $19.8\%$, having comparable $40.2\%$ object accuracy compared to OADis and CGE.

\paragrapht{VAW-CZSL.}
 We also achieve SoTA performance on VAW-CZSL, the most challenging benchmark having almost 10K composition labels for evaluation.
Even if we do not use additional unseen composition information like OADis, our method achieves the highest T@1 of $7.2\%$ with the best unseen label accuracy of $28.2\%$.
We also obtain both the highest attribute and object score (\tabref{tab:2}), especially with a significant improvement in attribute prediction by $2.8\%$ compared to CGE.
It proves that the careful modeling of the visual network in \method is essential to substantially improve the generalization capability. 
In addition, the results suggest that finetuning the visual backbone to suit composition recognition improves the performance in all benchmarks.

\subsection{Ablation study}
We ablate our CoT and MAA with different design choices on \tabref{tab:3}. We use the VAW-CZSL dataset and fine-tuned ViT-B backbone as our main ablation setting. 

\paragrapht{Component analysis.}
We report ablation results of \method and MAA in \tabref{tab:3a}.
In `without CoT' case, we use CompCos~\cite{mancini2021open} baseline. 
CoT boosts the AUC and HM significantly about 1.1$\%$ and 2.3$\%$ compared to the baseline. 
MAA offers better results to both \method and baseline, showing scalability to other methods.
Notably, employing MAA on CoT gives higher gains of 0.3$\%$ AUC and 2.3$\%$ HM compared to the baseline, demonstrating that the two components create a synergy effect. 

\paragrapht{Impact of loss functions.}
\tabref{tab:3b} shows the impact of each loss function. 
Compared to the case using $\mathcal{L}_{\text{comp}}$ only, $\mathcal{L}_{\text{obj}}$ and $\mathcal{L}_{\text{attr}}$ bring a gain on both AUC and HM.
Using both loses to $\mathcal{L}_{\text{comp}}$ improves AUC and HM significantly by $1.26\%$ and $2.8\%$ respectively, showing the importance of object and attribute losses to learn each expert.

\paragrapht{Different data augmentations.}
In \tabref{tab:3c}, we compare MAA with various data mixing methodologies including Manifold Mixup~\cite{verma2019manifold}, CutMix~\cite{yun2019cutmix} and Mixup~\cite{zhang2017mixup}.
To apply Manifold Mixup in \method, we select 3,6,9 and the last layer as a mixup layer, and mix these intermediate features.
MAA achieves the highest score on both AUC and HM, outperforming other augmentation methods including standard Mixup.
This demonstrates the effectiveness of MAA which balances the data distribution with a composition pair. 
Interestingly, we found that Manifold Mixup performs at worst, even degrading pure CoT performance (AUC: 6.93, HM: 20.4) in \tabref{tab:3a}.
We speculate that the mixed representation at intermediate layers can lead to significant underfitting of the attention module.

\paragrapht{Impact of object guidance.}
As shown in \tabref{tab:3d}, the object guidance significantly boosts the performance.
We found that reversing the guidance (attribute guidance) decreases the performance of AUC by 0.3$\%$ and HM by 0.7$\%$.
The results validate the usage of object guidance for contextualized attribute prediction.

\paragrapht{Design choice of feature emsemble.}
\tabref{tab:3e} shows the ablations with different configurations of intermediate blocks used in the attribute expert.
We evaluate four types of block feature ensembles: low, mid, high, and mixture.
Among them, the mixture type achieves the best performance on AUC, harmonic mean, and attribute accuracy. 
We therefore choose this mixture ensemble as a default setting.
Another finding is low-level feature is more adequate for attribute and unseen composition recognition, while high-level feature is robust to object and seen composition accuracy.

\subsection{Discussion}

\begin{figure}[!t]
\centering
{\includegraphics[width=0.99\linewidth]{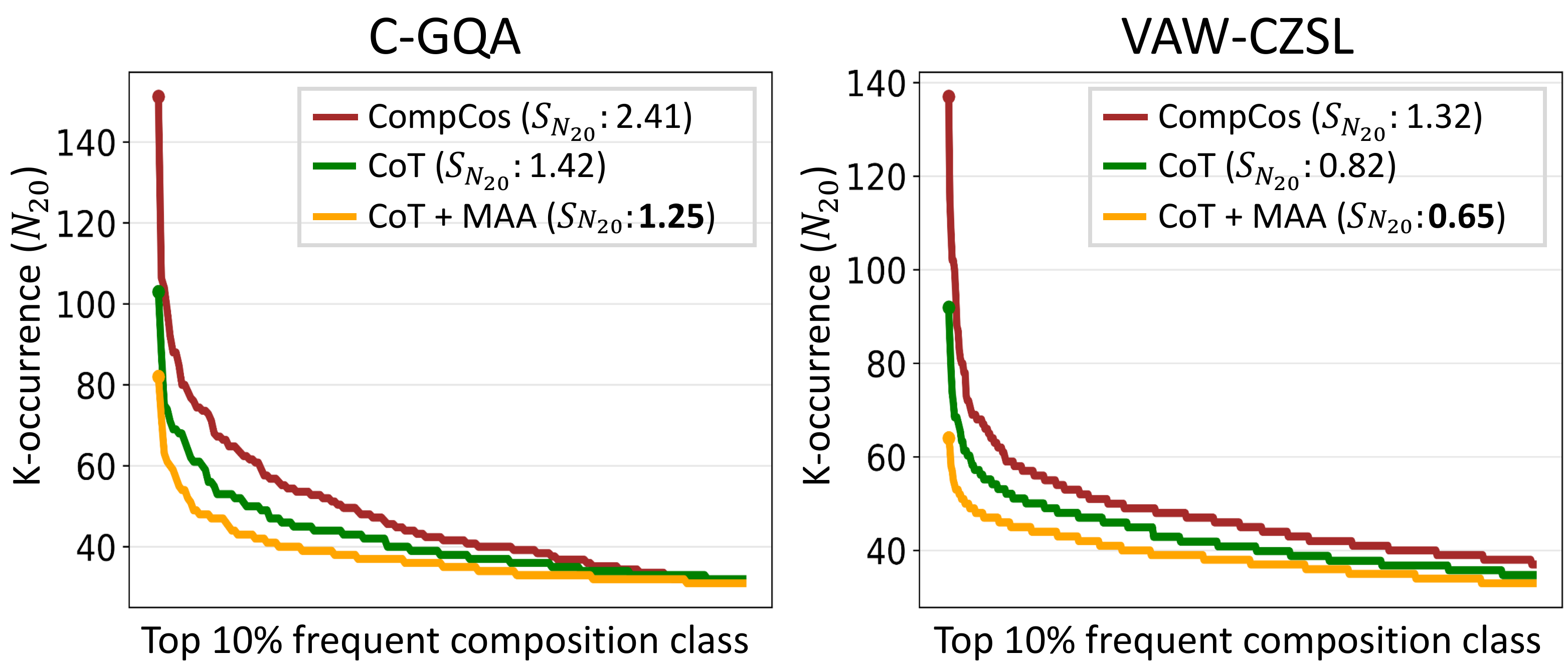}}\\
\caption{Distribution of k-occurrence counts with k=20 of C-GQA and VAW-CZSL test set. All results are reported with ViT-B backbone. x-axis stands for composition class ids which are ordered by decreasing count numbers. For better visualization, we show the top 10$\%$ head composition classes where the hubness is predominant.
Circles at start points indicate the most frequent hubs.
The result on MIT-States is in the Appendix C.1.
}
\label{fig:4}

\end{figure}

\begin{figure}[!t] 
\centering
        \includegraphics[width=0.99\linewidth]{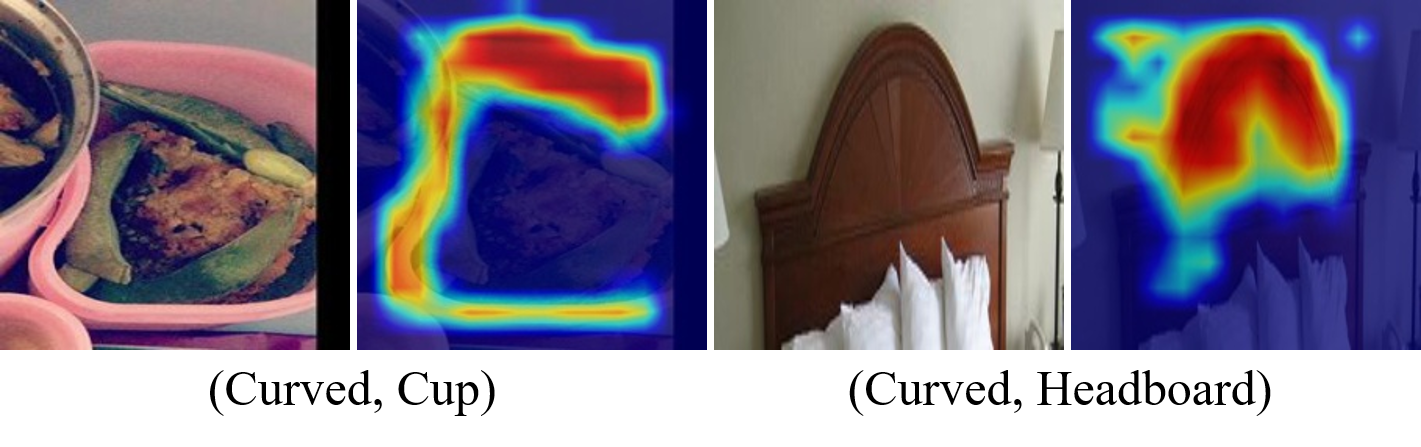} \\    
        \includegraphics[width=0.99\linewidth]{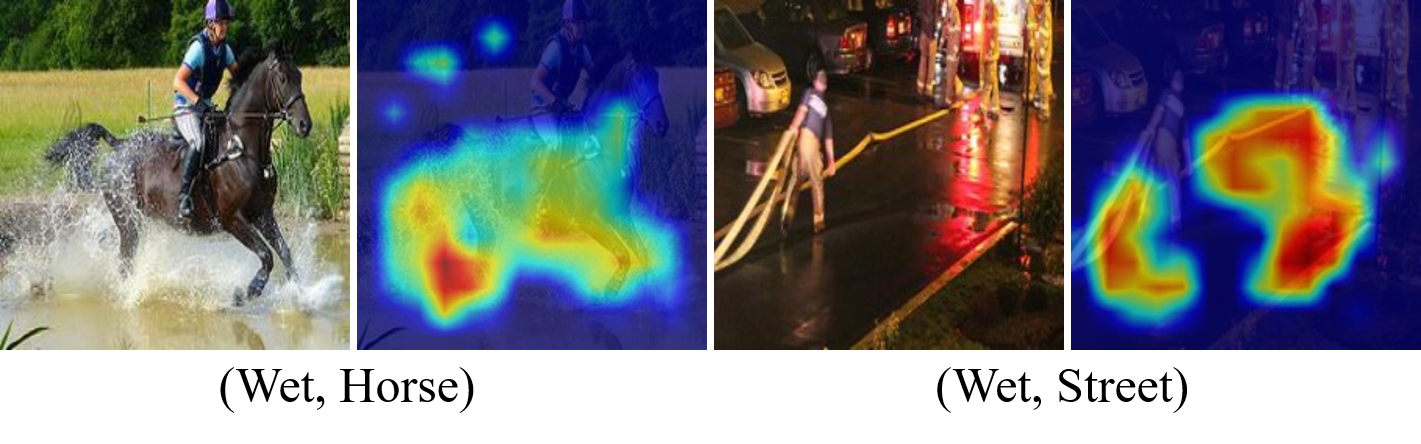}
        \vspace{-10pt}
    \caption{Visualization of object-guided attention on VAW-CZSL dataset. Each row represents the attended regions in image pairs, which are related to different attributes contextualized with a specific object. More results in the Appendix C.5.}
    \label{fig:5}
        
\end{figure}

\paragrapht{Mitigating hubness problem.}
We illustrate the hubness effect of a baseline~\cite{mancini2021open} and our method on C-GQA and VAW-CZSL datasets in \figref{fig:4}.
Concretely, we count k-occurrences ($N_k$), the number of times that a sample appears in the k nearest neighbors of other points~\cite{radovanovic2010hubs} for all class ids, and calculate the skewness of $N_k$ distribution for hubness measurement.
We observe that CoT significantly alleviates the hubness problem of baseline across datasets, enlarging the visual discrimination in the embedding space.
MAA further reduces the skewness of the distribution by mitigating biased prediction on head composition classes.

\paragrapht{Object-guided attention visualization.}
Qualitatively, we visualize attention maps from the object-guided attention module in \figref{fig:5}.
We observe that the attention module could capture the contextualized meaning of attributes in composition, \eg, wet skin of horse \vs puddle in street. 
It validates that the attention module assists \method to extract distinctive attribute embedding for composition by localizing specific RoI.

\begin{figure}[!t]
\centering
{\includegraphics[width=0.99\linewidth]{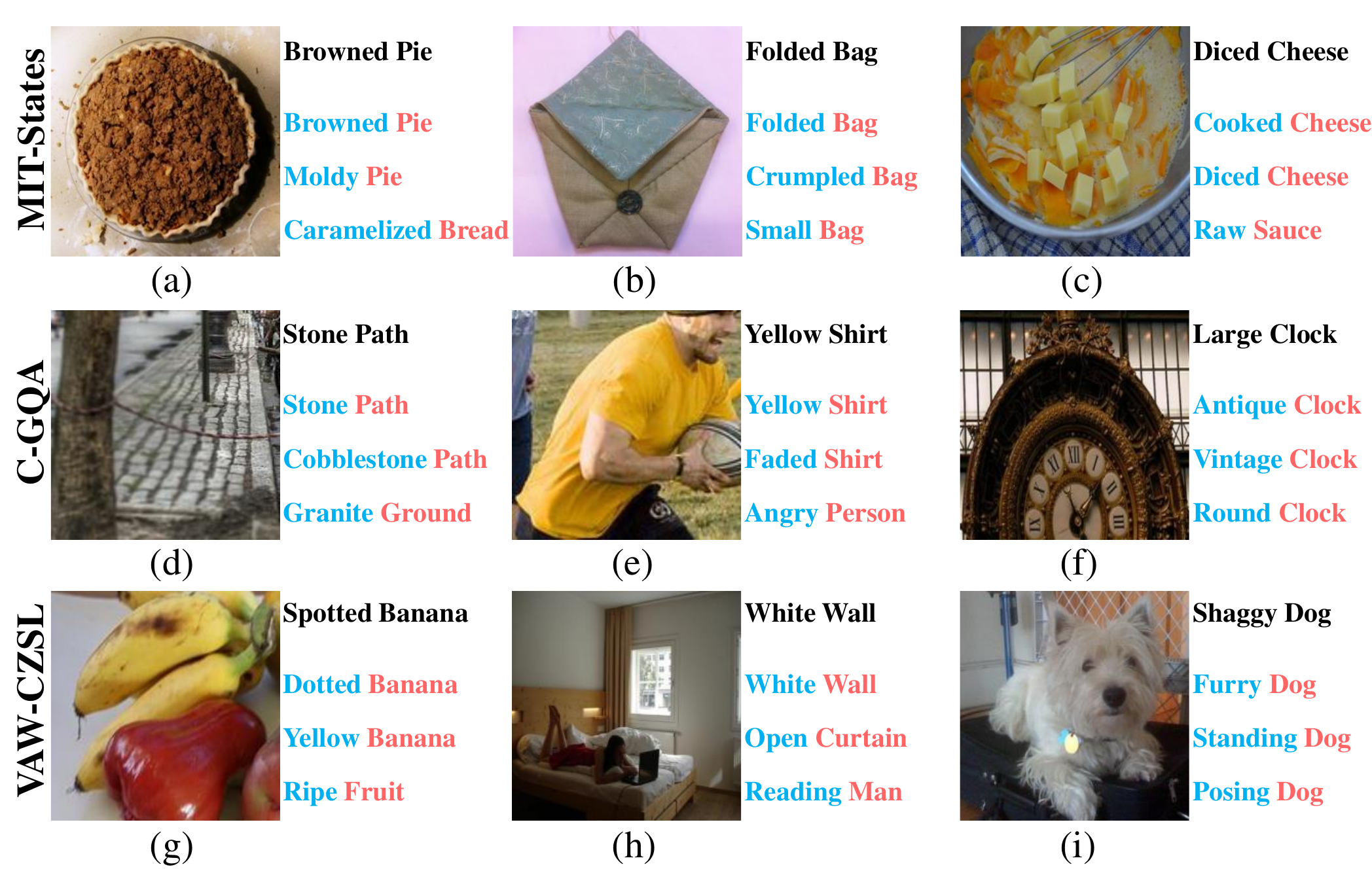}}\\\vspace{-3pt}
\caption{Ground-truth composition labels and top-3 prediction results on MIT-States, C-GQA and VAW-CZSL datasets. More results in the Appendix C.6.}
\label{fig:7}

\end{figure}

\paragrapht{Qualitative results.}
\figref{fig:7} illustrates the qualitative results of \method.
Even though some predictions do not correspond to annotated image labels, they explain the composition in another view; \eg, `Furry Dog' and `Shaggy Dog' in \figref{fig:7} (i).
Moreover, in-the-wild images usually have multiple objects with their own attributes; \eg, `White Wall' and `Reading Man' in \figref{fig:7} (h).
To recognize such compositions in the real world, it is necessary to handle the multi-label nature of composition, while distinguishing each attribute-object interaction \cite{hou2020visual} in an image.
It encourages us to construct a new CZSL benchmark along with a novel evaluation metric handling both multi-label prediction and multiple compositions in the image. 
We leave it as a future work.

\section{Conclusion}\vspace{-3pt}
In this paper, we propose Composition Transformer (CoT) with simple minority attribute augmentation (MAA), a contextual, discriminative, and unbiased CZSL framework.
The proposed object and attribute experts hierarchically utilize the entire powerful visual backbone to generate a composition embedding.
The simple but effective MAA balances the long-tail distributed composition labels.
Extensive studies with comprehensive analysis demonstrate the effectiveness of each component, and show that CoT surpasses existing methods across several benchmarks.

\paragraph{Acknowledgement.}
This research was supported by the Yonsei Signature Research Cluster Program of 2022 (2022-22-0002) and the KIST Institutional Program (Project No.2E31051-21-203).

{\small
\bibliographystyle{ieee_fullname}
\bibliography{egbib}
}

\newpage
\section*{Appendices}
\appendix

\section{Data statistics}\label{sec:a1}
\tabref{tab:1s} shows detailed data statistics of MIT-States~\cite{isola2015discovering}, C-GQA~\cite{naeem2021learning} and VAW-CZSL~\cite{saini2022disentangling}.
Compared to MIT-States, the latest C-GQA and VAW-CZSL have a large number of attribute, object and composition labels, effective for discussing CZSL problems on realistic scenarios.

\subsection{Long-tailed distribution}
\begin{figure*}[!t]
  \begin{subfigure}{0.33\linewidth}
    \includegraphics[width=\linewidth]{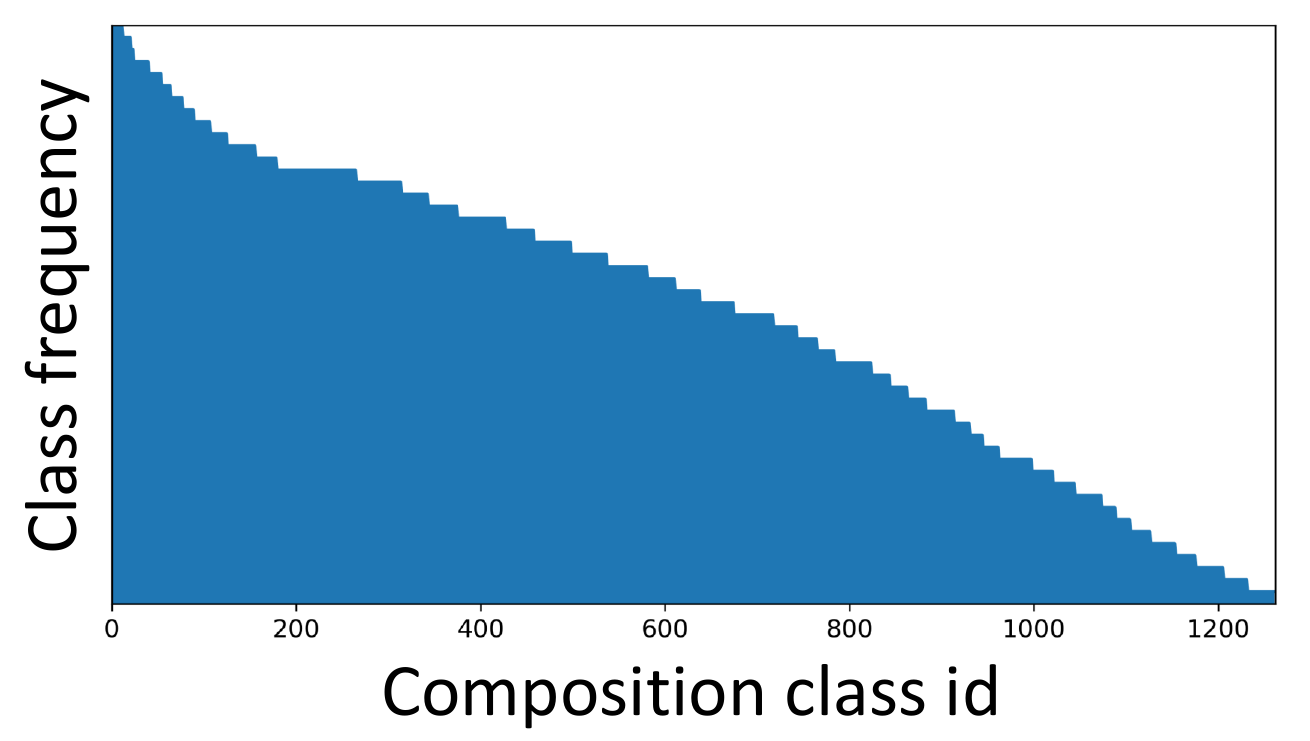}
    \caption{MIT-States} \label{fig:1a}
  \end{subfigure}%
  \hfill
  \begin{subfigure}{0.33\textwidth}
    \includegraphics[width=\linewidth]{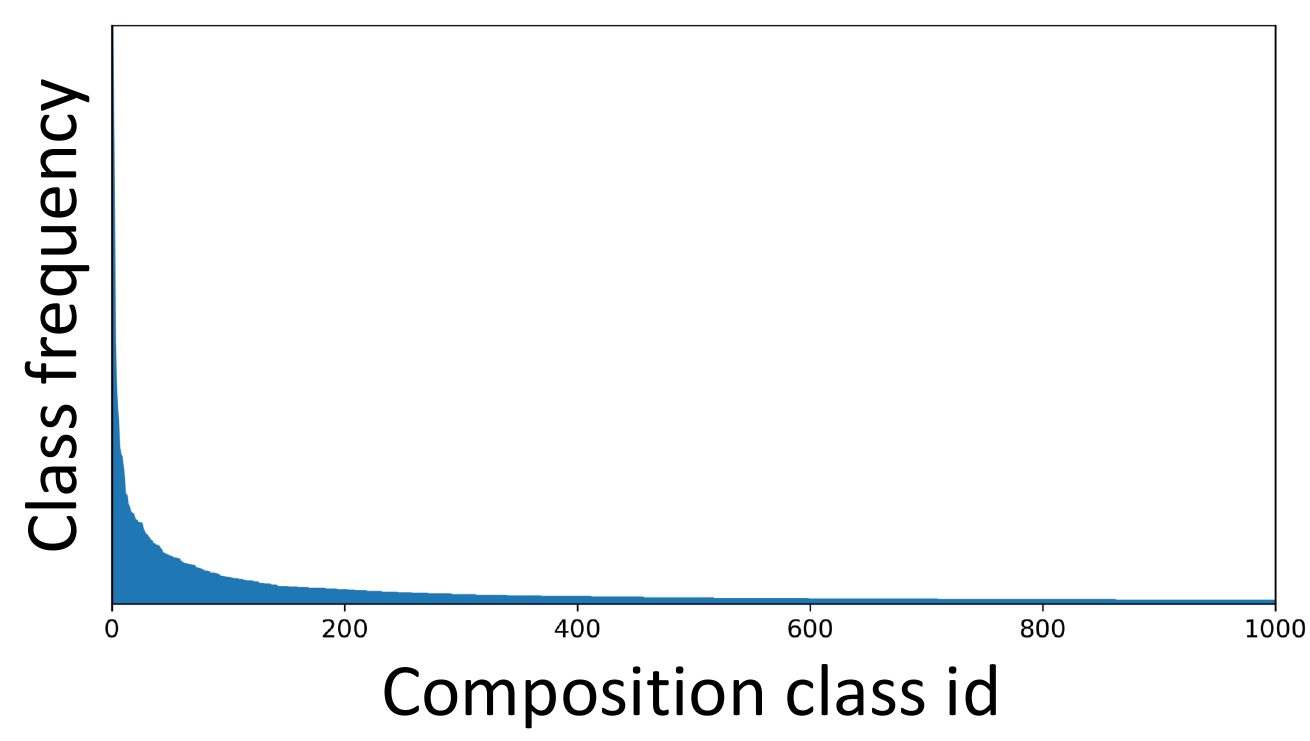}
    \caption{C-GQA} \label{fig:1b}
  \end{subfigure}%
  \hfill
  \begin{subfigure}{0.33\textwidth}
    \includegraphics[width=\linewidth]{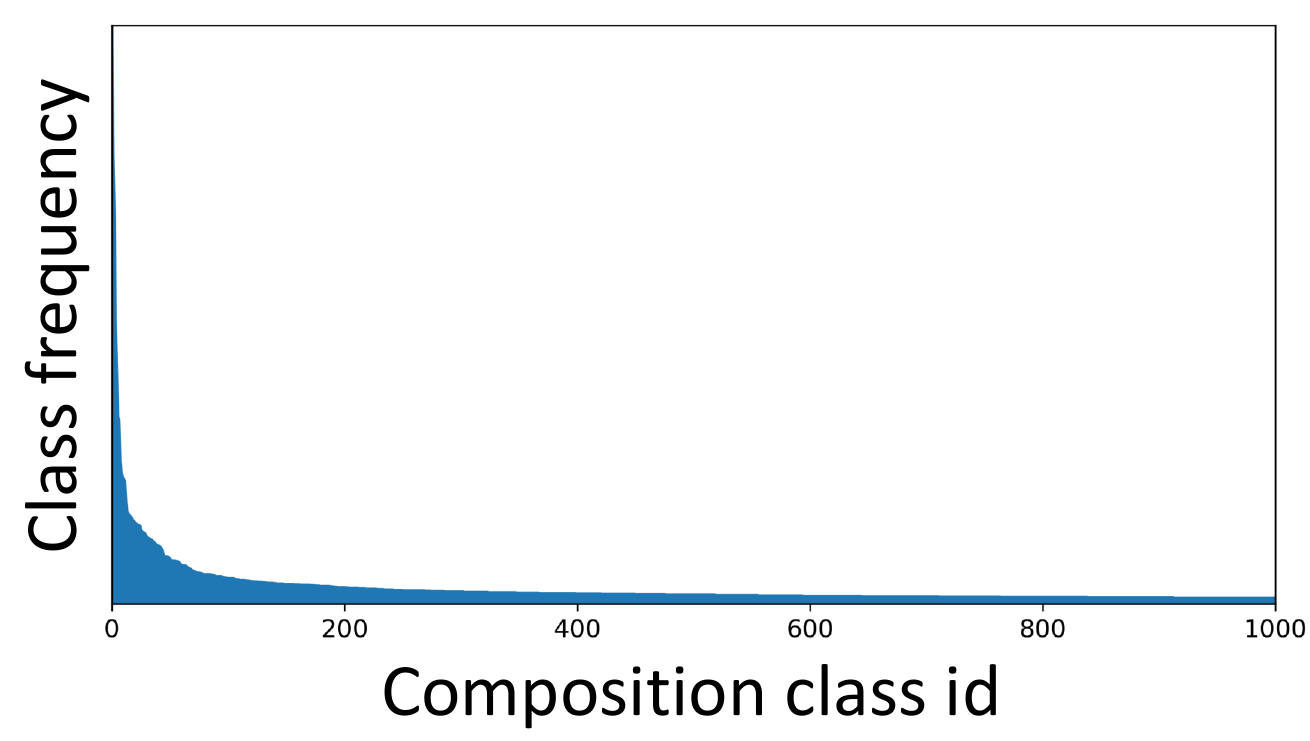}
    \caption{VAW-CZSL} \label{fig:1c}
  \end{subfigure}
  \hfill \\
    \vspace{-20pt}
\caption{Composition class distribution on three datasets. x-axis (composition class id) is ordered by decreasing composition frequency. For better visualization, we plot the top 1000 frequent composition classes on (b) and (c).} \label{fig:s1}
\end{figure*}

\begin{figure*}[tp!]
  \begin{subfigure}[t]{0.49\linewidth}
    \centering
    \includegraphics[width=0.49\linewidth]{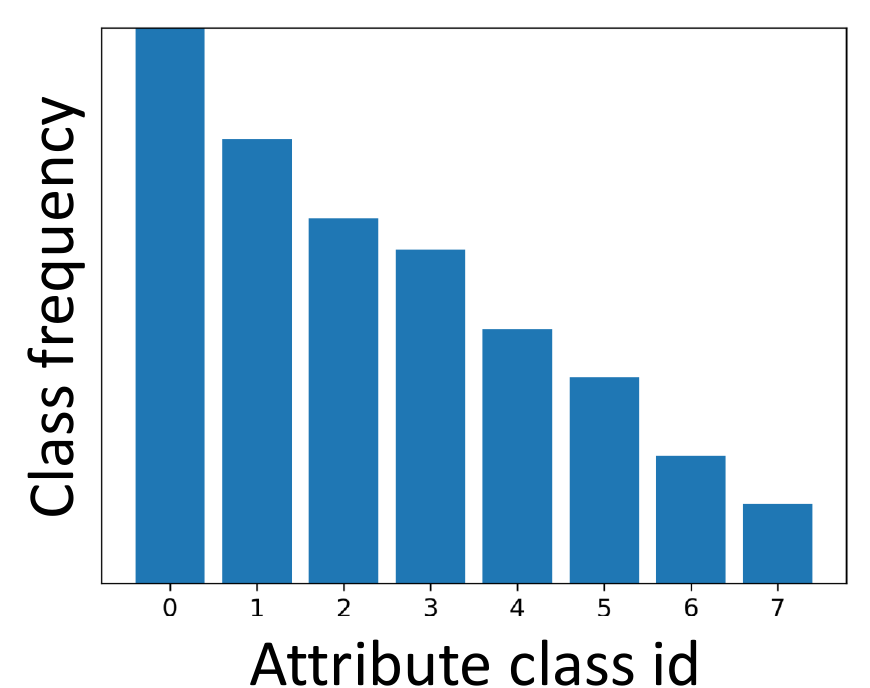}
    \includegraphics[width=0.49\linewidth]{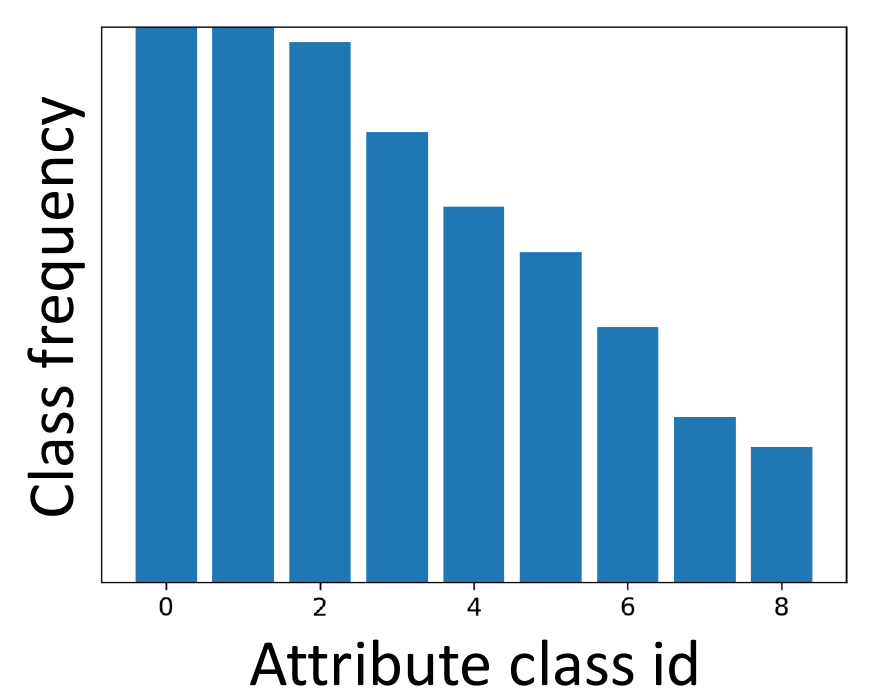}
    \caption{MIT-States}\label{fig:1a}
  \end{subfigure}
\hfill
  \begin{subfigure}[t]{.49\linewidth}
    \centering
    \includegraphics[width=0.49\linewidth]{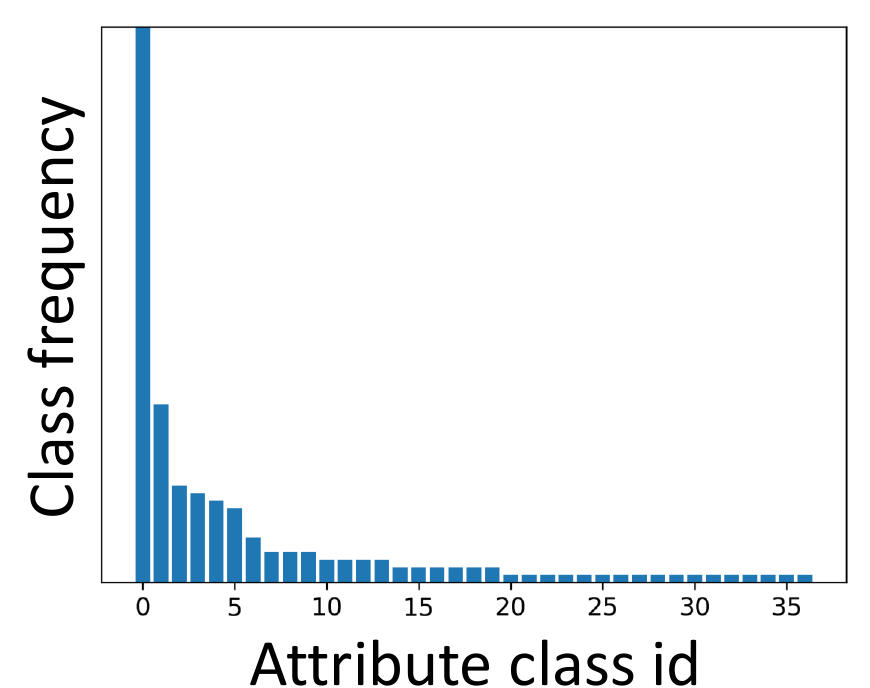}
    \includegraphics[width=0.49\linewidth]{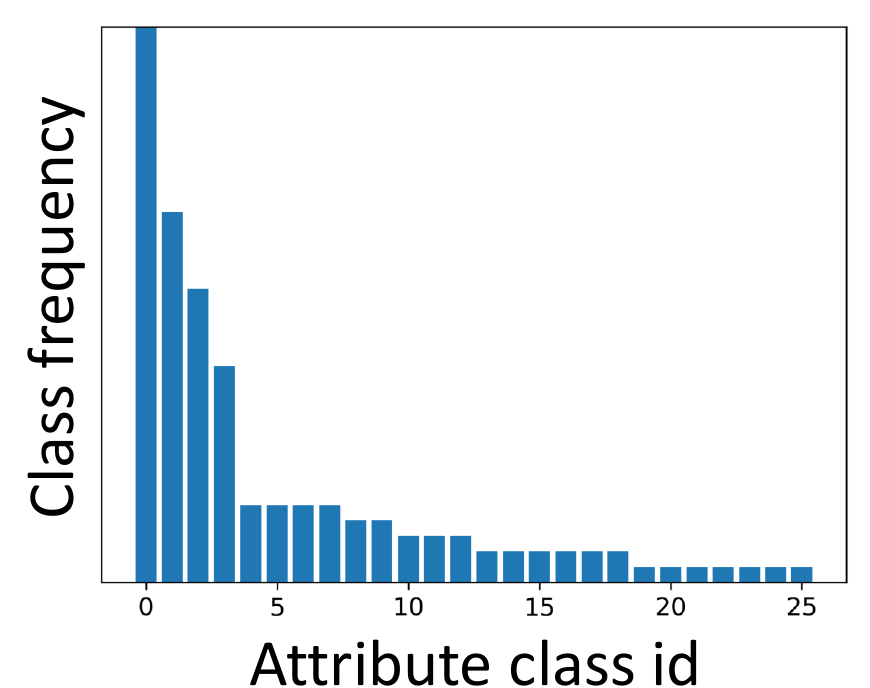}   
 \caption{C-GQA}\label{fig:1a}
  \end{subfigure}
        \vspace{-10pt}
  \caption{Attribute class distributions on MIT-States and C-GQA. Every attributes in each plot are composed into a fixed object; `Silk', `Bucket', `Street', and `Box'. (from left to right). }\label{fig:s2}
\end{figure*}

In \figref{fig:s1}, we visualize the distributions of composition class ids in the training set.
All datasets, especially for C-GQA and VAW-CZSL having a large number of composition classes, show the long-tailed distribution of compositions.
This is a natural effect because we can easily guess that `black dog' is more frequent than `blue dog' in the real world.
\figref{fig:s2} illustrates the imbalanced attribute composition (\eg, `white box' are 8 times more frequent than `pink box').
However, this phenomenon makes it difficult to predict various and novel compositions.
Therefore, we proposed minority attribute augmentation (MAA), which remedies a biased prediction caused by the imbalanced data distribution.

\section{Implementation details of MAA}\label{sec:a1}
We summarize our training procedure of the proposed MAA in \algref{alg:1}.
An auxiliary image $x_{B}$ is sampled with a sampling weight $\kappa$, which has a different attribute class to a given input $x_{A}$.
Then, a virtual sample $(x_{M}, y_{M})$ is generated by blending the input with the sampled auxiliary image.
We first optimize the object and attribute experts in \method with the generated virtual samples utilizing object and attribute losses (\ie, $\mathcal{L}_{\text{obj}}$ and $\mathcal{L}_{\text{att}}$) in the main paper.
To align a virtual visual prototype $\bp_{\textrm{vx}}$ from $x_{M}$ with a semantic prototype $\bp_{\textrm{vy}}=g([w(o), w(a_{M})])$, we simply modify the compositional loss $\mathcal{L}_{\text{comp}}$ with the virtual label as follows:
\begin{equation}
  \mathcal{L}_{\text{comp}}  =- \log \frac{\exp \{  d(\bp_{\textrm{vx}},\bp_{\textrm{vy}}) /\tau_c  \} }{\sum_{y_{k} \in \mathcal{Y}_{M}} \exp \{ d(\bp_{\textrm{vx}},\bp_{y_k})) /\tau_c  \} },
  \label{eq:loss}
\end{equation}
where $\mathcal{Y}_{M} = \mathcal{Y}_{s} \cup \{ y_{M} \}$.
We empirically find that directly applying the augmentation at the beginning of training leads to under-fitting.
To remedy this, the probability of applying MAA in each training iteration is gradually increased as training goes~\cite{yun2019cutmix, ghiasi2018dropblock}.
Specifically, we fix the initial probability as $0.3$ until the fifth epoch, while we linearly increase the probability by adding a factor of $0.1$ at every fifth epoch.

\begin{algorithm}
\caption{Minority attribute augmentation}\label{alg:1}
	\textbf{Require}: Training dataset $\mathcal{D}_{\text{tr}}$. \\
	\textbf{Initialize}: Model parameter $\theta$. 
 \medskip

\While{\text{Training}}{
  \For{$(x_{A},a_{A},o)$ $\in$ $\mathcal{D}_{tr}$}{
	        \While{$a_{B}$ $\neq$ $a_{A}$}
	        {Sample $(x_{B},a_{B},o)$ $\in$ $\mathcal{D}_{\text{tr}}$ \text{with} $\kappa(a_{B},o)$  
	        }
          Get $\bp_{x_{A}}$ and $\bp_{x_{B}}$ from \method 
          
          Sample   $\lambda \sim \textit{Beta}(1,1)$ 
             
         $\bp_{x_{M}}=\lambda\bp_{x_{A}}+(1-\lambda)\bp_{x_{B}}$
         
         $w(a_{M}) = \lambda w(a_{A}) + (1 - \lambda ) w(a_{B})$
         $\theta \leftarrow \theta - \nabla \mathcal{L}_{\text{total}} (\bp_{x_{M}}, w(a_M), w(o))$ 
}
}
\end{algorithm}

\begin{table*}[t]
    \small
    \centering
    \begin{tabular}{lccccccc}
\toprule
\multirow{2}{*}{Dataset} & \multirow{2}{*}{\#Attribute / \#Object} 
 & \multicolumn{2}{c}{Train} & \multicolumn{2}{c}{Validation} & \multicolumn{2}{c}{Test} \\ 
 \cmidrule(lr){3-4}
 \cmidrule(lr){5-6}
 \cmidrule(lr){7-8}
&   & $\mathcal{Y}_s$ & $\#$img & $\mathcal{Y}_s$ / $\mathcal{Y}_u$ & $\#$img   & $\mathcal{Y}_s$ / $\mathcal{Y}_u$ & $\#$img \\ \midrule

MIT-States\cite{isola2015discovering}       & 115 / 245  & 1262  & 30338  & 300 / 300 & 10420  & 400  / 400 & 12995 \\
C-GQA\cite{naeem2021learning}      & 453  / 870  & 6963  & 26920  & 1173 / 1368 & 7280  & 1022  / 1047 & 5098 \\
VAW-CZSL\cite{saini2022disentangling}     & 440  / 541  & 11175  & 72203  & 2121 / 2322 & 9524  & 2449  / 2470 & 10856 \\
\bottomrule
\end{tabular}\vspace{-5pt}
\caption{The statistics of three benchmarks: MIT-States\cite{isola2015discovering}, C-GQA\cite{naeem2021learning} and VAW-CZSL\cite{saini2022disentangling}.
}\label{tab:1s} 
  \vspace{-10pt}
\end{table*}

\section{More results}\label{sec:a2}

We provide more complementary results to validate \method on MIT-States, CGQA and VAL-CZSL benchmarks.

\begin{table}[t!]
    \begin{subtable}[h]{0.23\textwidth}
        \centering
    \scalebox{0.83}{
    \begin{tabular}{lcccc}
    \toprule
    CoT & MAA &  AUC & HM \\ 
    \midrule
     & & 8.89 & 22.9 \\ 
     & \checkmark & 9.08 & 23.5 \\
    \checkmark &  & 10.26 & 25.2 \\
    \checkmark & \checkmark &  \bf{10.54} & \bf{25.8} \\
\bottomrule
\end{tabular}}
       \caption{MIT-States}
       \label{tab:week1}
    \end{subtable}
    \hfill
    \begin{subtable}[h]{0.23\textwidth}
        \centering
    \scalebox{0.83}{
    \begin{tabular}{lcccc}
    \toprule
    CoT & MAA &  AUC & HM \\ 
    \midrule
     & & 5.25 & 17.6 \\ 
     & \checkmark & 5.58 & 20.3 \\
    \checkmark &  & 7.07 & 21.2 \\
    \checkmark & \checkmark &  \bf{7.42} & \bf{22.1} \\
\bottomrule
\end{tabular}}
       \caption{C-GQA}
        \label{tab:week2}
     \end{subtable}\vspace{-5pt}
     \caption{Component analysis on MIT-States and C-GQA dataset.}
     \label{tab:temps}
\end{table}


\begin{table}[t!]
    \centering
    \scalebox{0.95}{
    \begin{tabular}{lcc}
    \toprule
     Frequency type
    & AUC &  HM  \\ 
    \midrule
    $1/(\zeta_{o_{i}})^{0.5}$ & 7.19 & 21.5    \\
    $1/(\zeta_{o_{i}})^2$ & 7.07 & 21.2  \\
    $1/(\zeta_{o_{i}})$ [default setting in paper] & \bf{7.20} & \bf{21.7}  \\
\bottomrule
\end{tabular}} \vspace{5pt}
\caption{Ablation study for different frequency types of MAA on VAW-CZSL.}\label{tab:freq}
\end{table}
\begin{table}[t]
    \centering
    \scalebox{0.90}{
    \begin{tabular}{lcccc}
    \toprule
     Methods 
    & S & U & AUC & HM   \\ 
    \midrule
    CompCos~\cite{mancini2021open} & 25.4 & \uline{10.0} & 8.9 & 1.6 \\
    CGE~\cite{naeem2021learning} & 32.4 & 5.1 & 6.0 & 1.0 \\
    KG-SP~\cite{karthik2022kg} & 28.4 & 7.5 & 7.4 & 1.3   \\
    Ours (\method) & \uline{28.8} & \bf{11.3} & \bf{9.5} & \bf{1.8}  \\
\bottomrule
\end{tabular}}\vspace{-5pt}
\caption{Open-world CZSL results on MIT-States. All methods use Resnet18~\cite{he2016deep} backbone with a fine-tunning setup.}\label{tab:6}
\end{table}

 

\subsection{Hubness effect}
\begin{figure}[t]
\centering
{\includegraphics[width=0.7\linewidth]{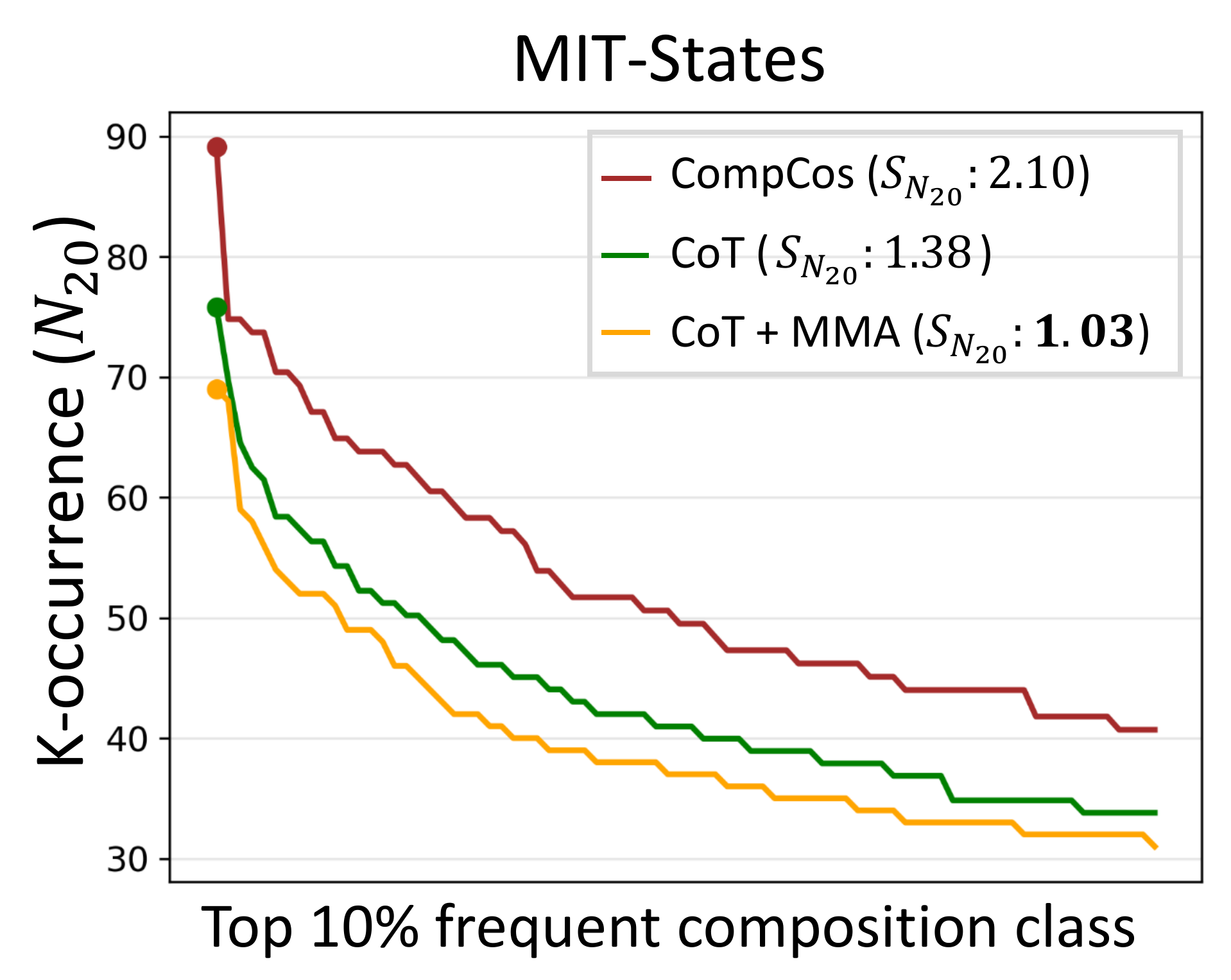}}\\\vspace{-10pt}
\caption{Distribution of k-occurence counts ($N_k$) of MIT-States test set. We use the same setting with Fig. 4 in main paper.}
\label{fig:s3}
\end{figure}

We illustrate a distribution of k-occurrences ($N_k$)~\cite{radovanovic2010hubs} to measure a hubness effect of visual features.
Note that different from \cite{dinu2014improving}, we calculate the nearest neighbors among visual features (queries) to analyze a hubness problem on the \textbf{visual domain} in both \figref{fig:s3} and \figref{fig:4} in the main paper.
To be consistent, the proposed CoT and MAA significantly alleviate the hubness problem by enhancing visual discrimination.

\subsection{Component analysis}
In \tabref{tab:temps}, we show the impact of each component (CoT and MAA) for CZSL performance on MIT-States and C-GQA datasets.
The ablation results together with Table 3a in the main paper demonstrate that both CoT and MAA consistently give improvements in AUC and HM.

\subsection{Other sampling weights}
In \tabref{tab:freq}, we conduct an ablation study for three different sampling weights leveraging an inverse attribute frequency $1/(\zeta_{o_i})$.
Notably, the frequency of $1/(\zeta_{o_i})^{2}$ yields worse performance.
It is under-fitting because sampling few tail class samples with too high probabilities prevents learning with other majority classes.
Sampling with square-root frequency, $1/(\zeta_{o_i})^{0.5}$, improves the performance on AUC and HM, but slightly below the result with $1/(\zeta_{o_i})$.  
We will include the above discussion in the paper.

\subsection{Open World setting}
To further analyze the generalization performance, we evaluate our \method on Open World setting~\cite{mancini2021open} in \tabref{tab:6}.
Following~\cite{mancini2021open}, we compute the best seen (S), unseen (U) accuracies, area under curve (AUC) and the best harmonic mean (HM). 
Our method also performs well in Open world setting, outperforming previous state-of-the-art methods in all metrics except the best seen accuracy.
This result demonstrates that enlarging visual discrimination with context modeling could also mitigate unfeasible compositions~\cite{mancini2021open} from a large output space of Open World scenario.

\subsection{Object-guided attention maps}
\begin{figure*}[!t] 
\centering
    \begin{subfigure}[b]{\linewidth}
        \includegraphics[width=0.99\linewidth]{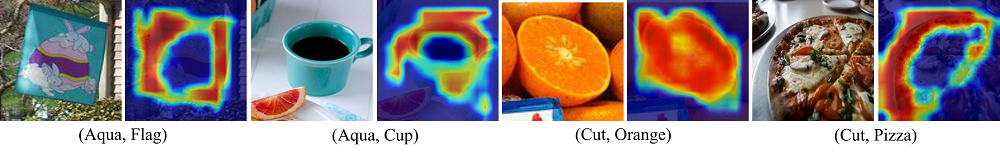} \vspace{-5pt}
    \end{subfigure}         
    \begin{subfigure}[b]{\linewidth}
        \includegraphics[width=0.99\linewidth]{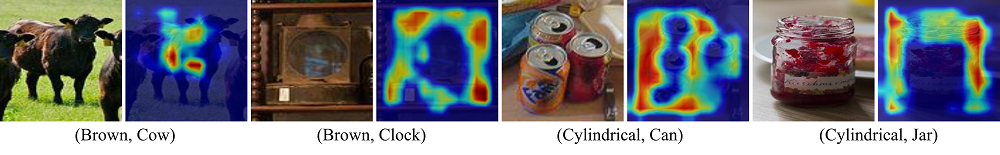} \vspace{-5pt}
    \end{subfigure} 
    \begin{subfigure}[b]{\linewidth}    
        \includegraphics[width=0.99\linewidth]{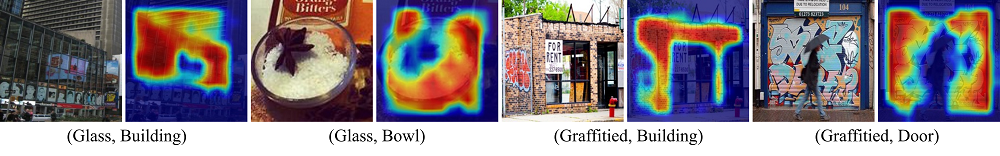} \vspace{-5pt}
    \end{subfigure} 
    \begin{subfigure}[b]{\linewidth}    
        \includegraphics[width=0.99\linewidth]{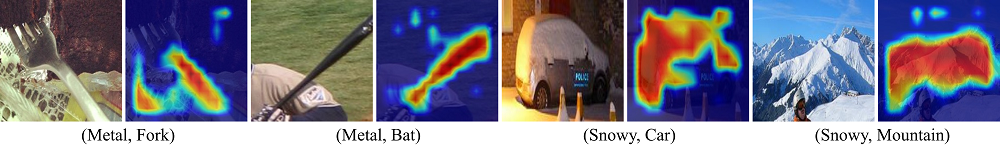} \vspace{-5pt}
    \end{subfigure} 
        \vspace{-20pt}
    \caption{Visualization of object-guided attention maps obtained on \textbf{VAW-CZSL}. 
    The input image and attended region by its specific attribute are paired with (attribute, object) labels. 
    (Attention weights: \textcolor{red}{High} to \textcolor{blue}{Low}).}
    \label{fig:s5}\vspace{-5pt}
\end{figure*}

\begin{figure*}[!t] 
\centering
    \begin{subfigure}[b]{\linewidth}    
        \includegraphics[width=0.99\linewidth]{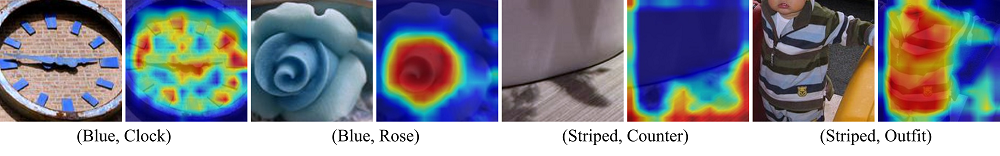} \vspace{-5pt}
    \end{subfigure} 
    \begin{subfigure}[b]{\linewidth}    
        \includegraphics[width=0.99\linewidth]{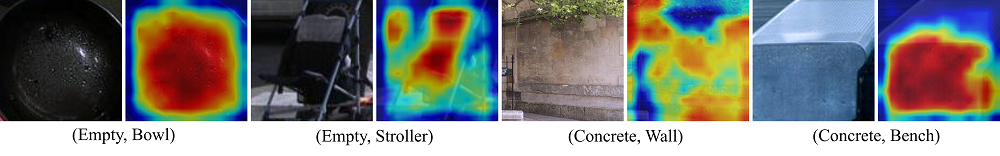} \vspace{-5pt}
    \end{subfigure} 
    \begin{subfigure}[b]{\linewidth}    
        \includegraphics[width=0.99\linewidth]{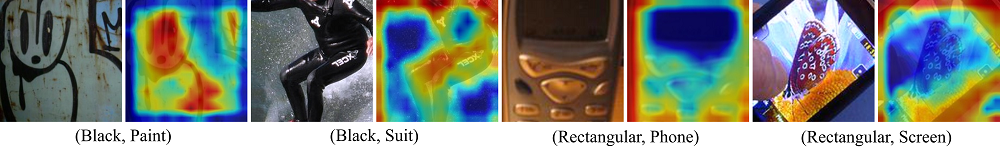} \vspace{-5pt}
    \end{subfigure} 
    \begin{subfigure}[b]{\linewidth}
        \includegraphics[width=0.99\linewidth]{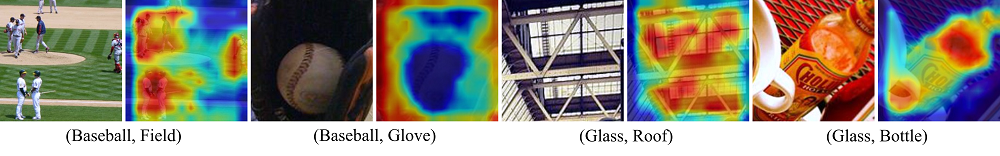} \vspace{-5pt}
    \end{subfigure} 
        \vspace{-20pt}
    \caption{Visualization of object-guided attention maps on \textbf{C-GQA}. 
    The input image and attended region by its specific attribute are paired with (attribute, object) labels. 
    (Attention weights: \textcolor{red}{High} to \textcolor{blue}{Low}).}
    \label{fig:s6}
\end{figure*}

We illustrate the object-guided attention maps in \figref{fig:s5} with VAW-CZSL, and \figref{fig:s6} with C-GQA.
For visualization, we merge three attention maps from low, middle and high blocks through multiplication.
The results demonstrate that the attention module could capture the contextualized regions for each attribute, enhancing the visual discrimination of attribute prototypes and its composition.

\subsection{Top-3 prediction results}
\begin{figure*}[t]
\centering
{\includegraphics[width=0.99\linewidth]{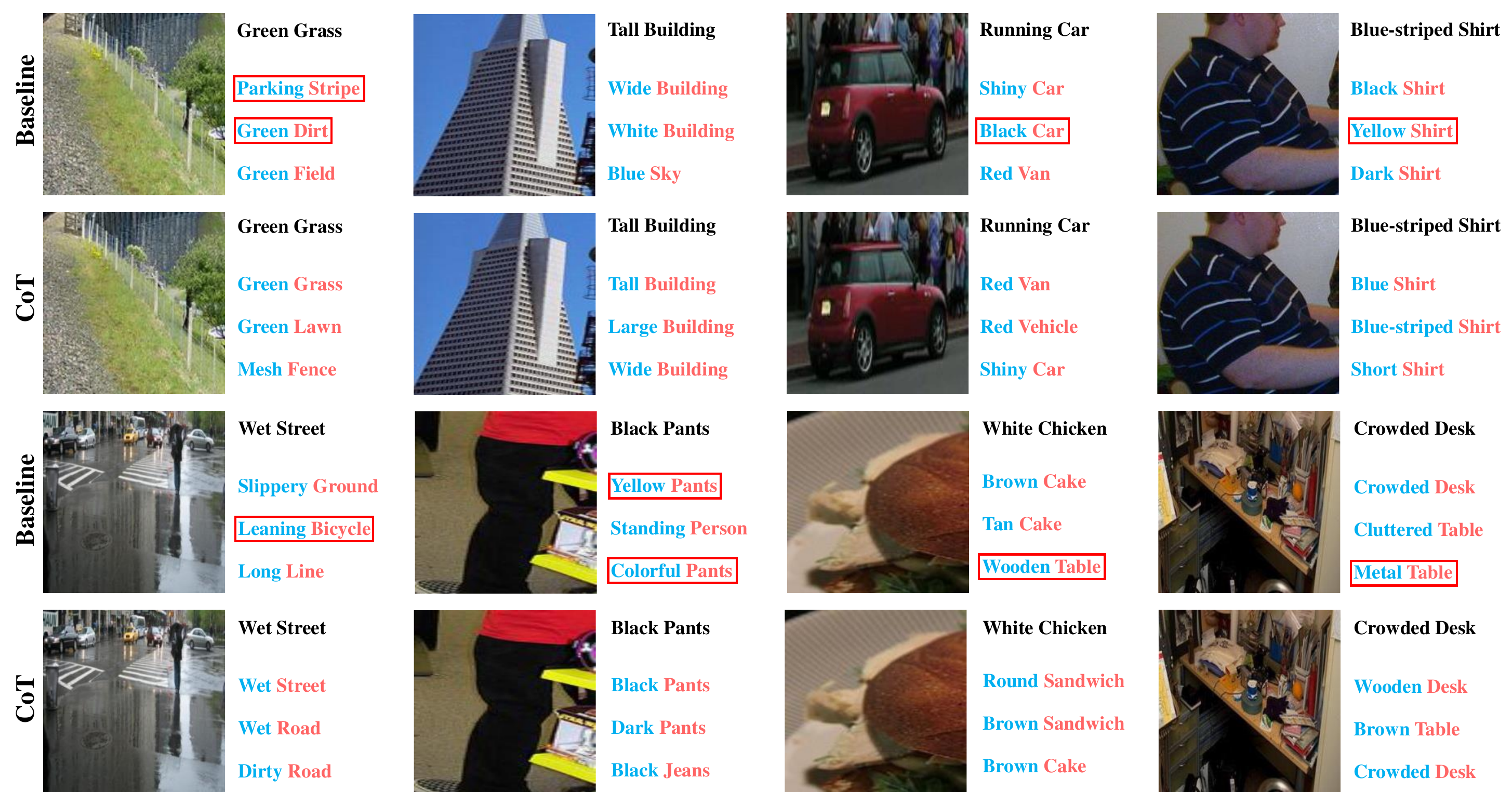}}\\
\caption{Ground-truth and top-3 prediction results on \textbf{VAW-CZSL}. We compare \method with baseline (OADis)~\cite{saini2022disentangling}. Red box denotes the false positive, having irrelevant or opposite compared to ground truth.}
\label{fig:s7}\vspace{-5pt}
\end{figure*}

\begin{figure*}[t]
\centering
{\includegraphics[width=0.99\linewidth]{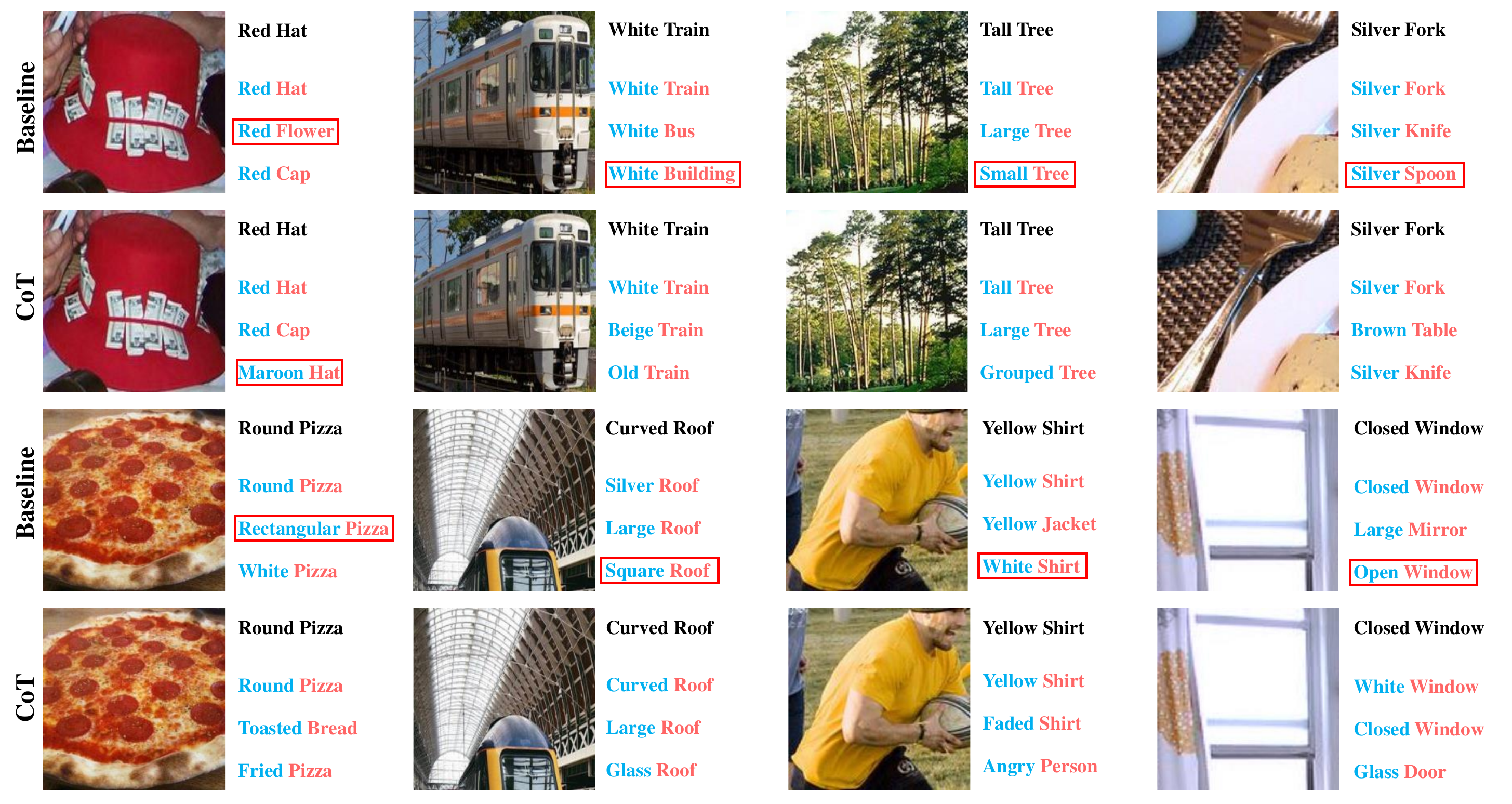}}\\
\caption{Ground-truth and top-3 prediction results on \textbf{C-GQA}. We compare \method with baseline (OADis)~\cite{saini2022disentangling}. Red box denotes the false positive, having irrelevant or opposite compared to ground truth.}
\label{fig:s8}\vspace{-5pt}
\end{figure*}

We visualize top-3 prediction results in \figref{fig:s7} with VAW-CZSL, and \figref{fig:s8} with C-GQA.
\method clearly outperforms the baseline~\cite{saini2022disentangling} to retrieve the relevant composition labels.
As discussed in Sec. 4.4 of the main paper, the qualitative results  show the limitation of existing CZSL datasets~\cite{naeem2021learning, saini2022disentangling} including multi-label composition and multiple attribute-object interaction. 

\section{Computational Analysis}\label{sec:a3}
In \tabref{tab:7}, we compare computational complexity with the previous state-of-the-art OADis~\cite{saini2022disentangling} by reporting the number of parameters and GFLOPs.
Although \method has more parameters induced from the ensemble of block features, it has almost the same model complexity in terms of GFLOPs, thanks to the parameter-efficient convolution based attention module.

\begin{table}[t]
    \small
    \centering
    \begin{tabular}{lcc}
    \toprule
   Methods & $\#$  Parameters (M)  & GFLOPs  \\ 
  \midrule
OADis\cite{saini2022disentangling}        & 2.25  & 11.2    \\

\method    & 3.34  & 11.8    \\
\bottomrule

\end{tabular}\vspace{-5pt}
\caption{Comparison of computational complexity between OADis~\cite{saini2022disentangling} and \method. Note that backbone (ViT-B) is excluded to count the parameters. }\label{tab:7}\vspace{-5pt}
\end{table}

\newpage

\end{document}